\documentclass[sn-mathphys-num]{sn-jnl}% Math and Physical Sciences Numbered Reference Style 
\usepackage{graphicx}%
\usepackage{multirow}%
\usepackage{amsmath,amssymb,amsfonts}%
\usepackage{amsthm}%
\usepackage{mathrsfs}%
\usepackage[title]{appendix}%
\usepackage{xcolor}%
\usepackage{textcomp}%
\usepackage{manyfoot}%
\usepackage{booktabs}%
\usepackage{algorithm}%
\usepackage{algorithmicx}%
\usepackage{algpseudocode}%
\usepackage{listings}%

\usepackage[leftcaption]{sidecap}
\usepackage[section]{placeins}
\usepackage{subcaption}

\newcommand{\red}[1]{\textcolor{red}{#1}}

\theoremstyle{thmstyleone}%
\theoremstyle{thmstyletwo}%

\theoremstyle{thmstylethree}%

\begin{document}

\title[SynMVCrowd: A Large Synthetic Benchmark for Multi-view Crowd Counting and Localization]{SynMVCrowd: A Large Synthetic Benchmark for Multi-view Crowd Counting and Localization}


\author[1]{\fnm{Qi} \sur{Zhang}}\email{qi.zhang.opt@gmail.com}

\author[1]{\fnm{Daijie} \sur{Chen}}\email{chendaijie2022@email.szu.edu.cn}

\author[1]{\fnm{Yunfei} \sur{Gong}}\email{gongyunfei2021@email.szu.edu.cn}

\author*[1]{\fnm{Hui} \sur{Huang}}\email{hhzhiyan@gmail.com}

\affil[1]{\orgdiv{Visual Computing Research Center, College of Computer Science and Software Engineering}, \orgname{Shenzhen University}, \orgaddress{ \city{Shenzhen}, \country{China}}}

%\affil[2]{\orgname{Guangdong Laboratory of Artificial Intelligence and Digital Economy (SZ)}, \orgaddress{ \city{Shenzhen}, \country{China}}}

% \affil[3]{\orgdiv{Department}, \orgname{Organization}, \orgaddress{\street{Street}, \city{City}, \postcode{610101}, \state{State}, \country{Country}}}


\abstract{Existing multi-view crowd counting and localization methods are evaluated under relatively small scenes with limited crowd numbers, camera views, and frames. This makes the evaluation and comparison of existing methods impractical, as small datasets are easily overfit by these methods. To avoid these issues, 3DROM \cite{qiu20223d} proposes a data augmentation method. Instead, in this paper, we propose a large synthetic benchmark, SynMVCrowd, for more practical evaluation and comparison of multi-view crowd counting and localization tasks. The SynMVCrowd benchmark consists of 50 synthetic scenes with a large number of multi-view frames and camera views and a much larger crowd number (up to $1000$), which is more suitable for large-scene multi-view crowd vision tasks. Besides, we propose strong multi-view crowd localization and counting baselines that outperform all comparison methods on the new SynMVCrowd benchmark. Moreover, we prove that better domain transferring multi-view and single-image counting performance could be achieved with the aid of the benchmark on novel new real scenes. As a result, the proposed benchmark could advance the research for multi-view and single-image crowd counting and localization to more practical applications. The codes and datasets are here: https://github.com/zqyq/SynMVCrowd.}




\maketitle

\section{Introduction}\label{sec1}

\label{sec:intro}

\begin{figure*}[htbp!]
\centering
\includegraphics[width=\linewidth]{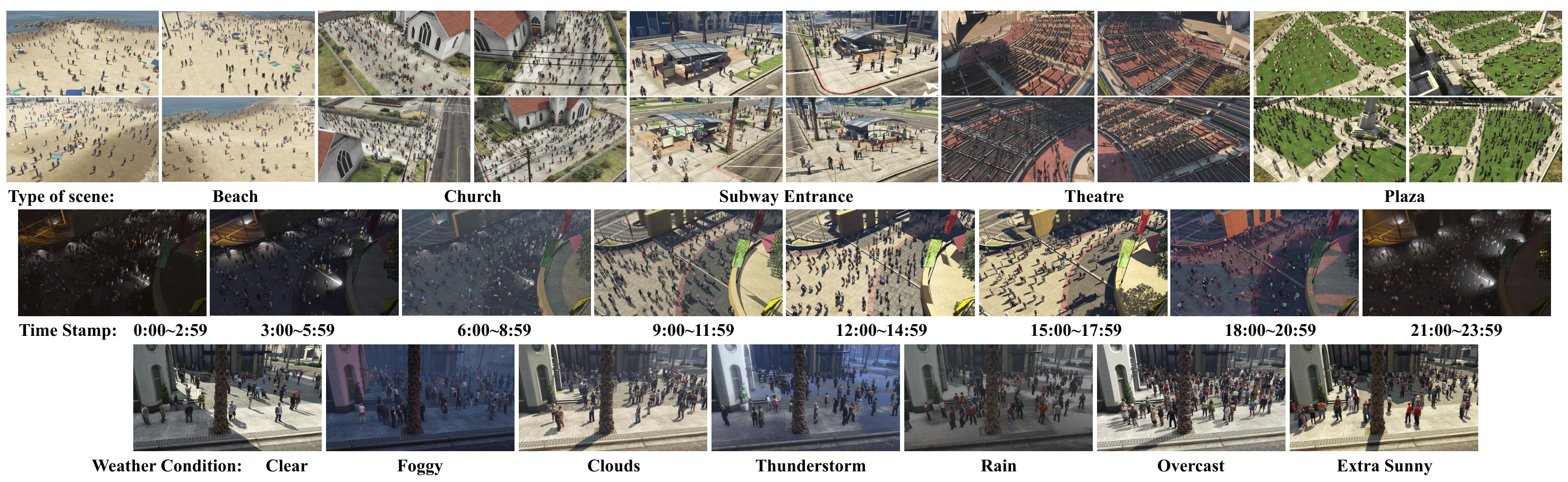}
\vspace{-0.5cm}
   \caption{The visualization of the proposed SynMVCrowd dataset: the type of scene, time stamps, and the weather condition.}
\vspace{-0.4cm}
\label{fig:display}
\end{figure*}

Crowd vision tasks include crowd counting \cite{zhang2016single}, crowd detection \cite{ren2015faster}, and human pose estimation \cite{andriluka20142d}, \textit{etc}.
To handle the severe occlusions under large scenes, multi-camera views are adopted for more challenging crowd vision tasks, such as multi-view crowd counting \cite{zhang2021CVCS}, multi-view crowd localization \cite{hou2020multiview}, or 3D pose estimation \cite{iskakov2019learnable}, \textit{etc}.
The existing multi-view crowd datasets \cite{chavdarova2018wildtrack,hou2020multiview,ferryman2009pets2009} are limited in scenes, views, frames, and crowd numbers, such as the Wildtrack dataset \cite{chavdarova2018wildtrack} for multi-view crowd localization, which only includes 400 frames, 7 camera views, and 1 scene with about 30 people. These limitations greatly reduce the applicability of the current multi-view crowd vision methods in the real world.

However, collecting and annotating real-world multi-view image datasets with hundreds of scenes, camera views, and frames, and thousands of crowds, is too expensive. Instead, generating synthetic data becomes an alternative solution. Previous research in the area has proposed several synthetic datasets for these problems, such as MultiviewX \cite{hou2020multiview}. However, MultiviewX is also a multi-view image dataset with limited camera views, scene numbers, and crowd numbers.
There are also a lot of single-image-based large synthetic datasets like GCC \cite{Wang2019Learning}, but they are not meant for multi-view settings.
CVCS dataset \cite{zhang2021CVCS} is a large synthetic dataset for multi-view counting tasks, which includes 31 scenes, with around 100 camera views and 100 frames per scene. However, the crowd number in the CVCS dataset is still too small ($<200$), which is not difficult enough for multi-view settings.
%Besides, the CVCS dataset is designed mainly for multi-view crowd counting, not considering other tasks in the dataset construction.

To solve the mentioned problems, we propose SynMVCrowd,
% (see Fig. \ref{fig:abstract_figure})
 a large synthetic benchmark for practical multi-view-based crowd vision tasks, such as multi-view crowd counting and multi-view crowd localization. 
The proposed synthetic benchmark SynMVCrowd includes 50 scenes, 50 camera views, and 200 frames per scene, with around 200-1000 people in each scene. 
SynMVCrowd is the largest multi-view crowd dataset for various crowd vision tasks, which shall advance the multi-view crowd counting and localization from single-scene settings to cross-scene cross-view settings. With so many images of various camera perspectives, weather conditions, and light variations and scenes  (see Fig. \ref{fig:display}), SynMVCrowd could also be utilized as a large and challenging benchmark for single-image crowd counting and localization tasks. 
We testify the state-of-the-art multi-view crowd counting and localization methods, such as \cite{hou2020multiview,song2021stacked,hou2021multiview,qiu20223d,zhang2024multi,zhang2021CVCS,mo2024countformer}, on the benchmark with the cross-scene setting, which could evaluate and compare existing methods under more practical conditions.
We also evaluate existing single-image crowd counting and localization methods, such as \cite{li2018csrnet,wang2020distribution,song2021rethinking}, on the proposed SynMVCrowd benchmark.
Overall, the proposed SynMVCrowd benchmark could advance the research for single-image and multi-view crowd counting and localization to more practical and challenging applications.
The contributions of the paper can be summarized as follows.

\begin{itemize}
  \item As far as we know, this is the largest synthetic dataset for multi-view and single-image crowd vision tasks, which consists of a large number of crowd scenes, camera views, and frames with a large crowd number per scene.
  
  \item We evaluate and compare the existing methods on SynMVCrowd, indicating it is a challenging platform for multi-view counting and localization under cross-scene settings. We propose strong multi-view counting and localization baselines that outperform all comparisons.
  Besides, SynMVCrowd could also serve as a benchmark for single-image counting and localization.
  
  \item We show that better multi-view and single-image domain transferring counting performance could be achieved with SynMVCrowd. SynMVCrowd provides new directions for research on multi-view and single-image counting and localization tasks.
\end{itemize}

\section{Related Work}
\label{sec:relatedwork}

\subsection{Multi-view Crowd Counting and Localization}
%Introduce multi-view crowd counting, multi-view people detection, and crowd segmentation. 

\textbf{Multi-view crowd counting.} 
%Given that occlusion is a challenging issue in single-image crowd counting tasks, researchers advocate the use of images captured from multiple cameras as input to address the occlusion issues.
Traditional multi-view counting methods \cite{Ge2010Crowd, li2012people, Maddalena2014people, Ryan2014Scene, Tang2014Cross} are typically trained on the PETS2009 dataset \cite{ferryman2009pets2009}, which contains only 2 to 4 views and a few hundred frames. These methods employ foreground extraction techniques and hand-crafted features, whose performance is limited. In recent years, approaches such as \cite{zhang2019wide, zhang20203d} have introduced end-to-end models that fuse cues from multiple views to enhance counting performance. \cite{zhang2019wide, zhang20203d} were trained on the CityStreet dataset \cite{zhang2019wide},  despite its large-scale scenes, which are constrained by limited crowd density, camera numbers, and frame counts. Another notable contribution is \cite{zhang2021CVCS}, which presents a cross-view cross-scene multi-view crowd counting method and proposes a large synthetic dataset CVCS. 
\cite{zhang2022single} and \cite{zhang2022calibration, li2025wscf} tried to reduce the requirement for multi-cameras and deal with the unsynchronized and uncalibrated multi-view counting tasks, respectively.
Semi-MVCC \cite{zhang2025semi} proposed a multi-view ranking-based model for semi-supervised multi-view counting.
CountFormer \cite{mo2024countformer} adopted a transformer-based architecture for multi-view crowd counting instead of CNN-based architectures and achieved new SOTA performance.
Although CVCS and CountFormer employ a synthetic dataset larger than their predecessors, it still falls short in applicability to highly congested scenes due to its restricted crowd density (with an average number of 135 in each scene).
%In CVCS, training and testing occur across different scenes with arbitrary camera layouts. Although CVCS employs a synthetic dataset larger than its predecessors, it still falls short in applicability to highly congested scenes due to its restricted crowd density (with an average of 135). To address this limitation, our proposal involves the creation of a synthetic dataset specifically tailored for highly congested scenes.
%\zqnote{add new refs later.}

\textbf{Multi-view crowd localization.}
In contrast to crowd counting, multi-view crowd localization emphasizes the precision of location detection. 
Recent methodologies \cite{baque2017deep, chavdarova2017deep, Aung2024} have shifted towards projecting feature maps onto the ground plane to fuse information from multiple RGB cameras, departing from the conventional approaches of fusing information for multi-view 2D anchors.
MVDet \cite{hou2020multiview} proposed a network that extracts multi-view feature maps and projects them onto the ground plane, then fuses them on the ground plane to obtain the final crowd location prediction.
MVDeTr \cite{hou2021multiview} incorporated attention modules to ensure optimal weight assignment to each perspective during information fusion. Building upon MVDet, SHOT \cite{song2021stacked} employs a soft-selection module for fusing cues from projection planes of different planes. 
3DROM \cite{qiu20223d} dealt with the overfitting issue through a 3D cylinder data augmentation technique.
SVCW \cite{zhang2024multi} proposed a large scene multi-view crowd localization model by introducing extra single-view ground-plane supervision via view contribution weighting in the fusion process.
MVOT \cite{zhang2024mahalanobis} proposed a novel point-supervision-based optimal transport for the multi-view crowd localization task, achieving much better performance than the Gaussian density map MSE loss. TrackTacular \cite{Teepe_2024_CVPR} proposed to utilize a 3D projection to achieve more precise multi-view fusion features. CaMuViD \cite{Daryani_2025_CVPR} introduced a forward–backward projection technique that enables calibration-free multi-view crowd localization.

\textit{Overall, many of these methods are trained and tested on small-scale scene datasets with sparse crowds in local areas, which hinders their ability to generalize to novel scenes in real-world scenarios. Therefore, a large multi-view crowd dataset with larger scenes, higher crowd densities, increased scene numbers and camera numbers, and variable camera layouts, is in demand.}
%^Recognizing the disparities between life scenarios and algorithm implementation scenarios, there arises a necessity for a large-scale dataset characterized by larger scenes, higher crowd densities, increased camera numbers, and variable layouts.

\subsection{Multi-view Crowd Datasets}
% Introduce multi-view crowd datasets: real, synthetic.
% \zq{write according to real, synthetic dataset?}
%\hspace{1em}Upon reviewing pertinent literature \cite{ferryman2009pets2009,ristani2016performance,chavdarova2018wildtrack,hou2020multiview,zhang2019wide,zhang2021CVCS} concerning multi-view crowd datasets in recent years, we found many datasets that can be used for multi-view crowd counting, localization, and so on. 
%Existing multi-view crowd datasets \cite{ferryman2009pets2009,ristani2016performance,chavdarova2018wildtrack,hou2020multiview,zhang2019wide,zhang2021CVCS} (see Table \ref{table:Multi-view_datasets}) can be broadly categorized into two groups based on the data source: real scene datasets and synthetic scene datasets.

\cite{ferryman2009pets2009,ristani2016performance,chavdarova2018wildtrack,zhang2019wide} are single-scene multi-view crowd datasets captured in the real world. %\textbf{PETS2009} \cite{ferryman2009pets2009} is comprised of three sections featuring a sequence of multiple views capturing pedestrians in an outdoor environment and consists of 3 camera views with 20-40 people in the scene.
% %is comprised of three sections featuring a sequence of multiple views capturing pedestrians in an outdoor environment. These segments are utilized for tasks such as crowd counting and density estimation, crowd tracking, and flow analysis through event recognition.
% \textbf{DukeMTMC} \cite{ristani2016performance} is a surveillance dataset captured on a campus with 4 camera views and 10-30 people in the scene. 
% %It serves as a resource for the research and development of video tracking systems, person re-identification, and low-resolution face recognition.
% \textbf{Wildtrack} \cite{chavdarova2018wildtrack} is a high-resolution dataset comprising 400 multi-view frames, which is captured with 7 static cameras in a public open area with a size of $12$m$\times36$m.
\textbf{PETS2009} \cite{ferryman2009pets2009}, \textbf{DukeMTMC} \cite{ristani2016performance} and \textbf{Wildtrack} \cite{chavdarova2018wildtrack} contain relatively fewer frames and crowd numbers ($<40$), covering a small scene size (see Table \ref{table:Single-image_datasets}).
\textbf{CityStreet} \cite{zhang2019wide} is an urban traffic intersection dataset containing 500 frames with 3 camera views and a larger crowd number (70-150) in the scene. In comparison to previous datasets, CityStreet significantly improves the coverage of scenes within the camera's perspective, but it is still too small for more complicated scenarios. 
\textit{To reduce the data collecting and labeling cost, synthetic scene datasets are also adopted}.
% \textbf{MultiviewX} \cite{hou2020multiview} and \textbf{CVCS} \cite{zhang2021CVCS} are synthetic scene datasets, created through computer synthesis to emulate real-world scenarios. 
\textbf{MultiviewX} \cite{hou2020multiview} covers a 16$\times$25 meters area, with 6 fixed camera views, only 400 multi-view frames, and 40 people in the scene on average.
In addition, \textbf{CVCS} \cite{zhang2021CVCS} is a large dataset with 31 diverse scenes and 31,000 multi-view frames in total, averaging 60 to 120 camera views and 90-180 people per scene.
Notably, CVCS significantly surpasses MultiviewX in crowd density, scene size, and annotated image count.
We conclude that current multi-view datasets still contain a small number of crowds in the scene ($<200$), which is not suitable for more practical applications.
Thus, a more challenging multi-view dataset with more crowds is in demand.
%\zqnote{add new refs later.}

%MultiviewX, a multi-view synthetic dataset, tends to have smaller scene sizes. In contrast, CVCS introduces various scene types with larger sizes. Moreover, MultiviewX pictures average around 40 people, while CVCS exhibits an average crowd density of approximately 135 individuals per picture. Both datasets may have limitations in specific scenarios like stadiums, rallies, or concerts due to their restricted crowd density. 
% This underscores the need for a comprehensive dataset to address these challenges.

%\vspace{-0.3cm}
\subsection{Single Image Counting and Localization}
Early single-image methods \cite{hafeezallah2021u, xie2020rsanet, liu2020crowd, liu2019crowd, sindagi2019multi, Huang2020stacked,cheng2019improving, ma2022fusioncount, Chen2024} design multi-scale networks to address the scale variation issues.
%Scale variation also brings the problem of non-uniform crowd distribution, which makes the neural networks obtain a relatively high accurate counting in the sparse region but fail in the dense region. 
\cite{sindagi2017generating, sam2018divide, babu2017switching, liu2018decidenet, xu2019learn, xiong2019open, jiang2020attention} and CSRNet \cite{li2018csrnet} used the context information to alleviate the non-uniform crowd distribution.
Besides, \cite{jeong2022congestion, ma2019bayesian, zhang2022crossnet} have proposed different loss functions to improve the model's attention to high-density crowd areas for better counting performance. P2PNet \cite{song2021rethinking} and DM-Count \cite{wang2020distribution} utilize point supervision without applying Gaussian kernels to dot maps. %P2PNet utilizes the Hungarian algorithm for one-to-one matching between predictions and ground truth, employing nAP as the evaluation metric to enhance both localization and counting performance. %DM-Count, on the other hand, uses optimal transport to evaluate the similarity between normalized predictions and ground truth. 
GramFormer \cite{lin2024gramformer} improves crowd counting by using graph-modulated transformers to diversify attention and weight features by node importance.
FreeLunch \cite{meng2025free} boosts multi-modal crowd counting by adding cross-modal alignment and regional density supervision with no extra data, parameters, or inference cost.
Recently, some researchers have turned to semi-supervised and weakly supervised methods \cite{you2023few, peng2022semi, lei2021towards, meng2021spatial} to make the model perform better in the absence of data.
%\zqnote{add new refs later.}

Existing single-image datasets include ShanghaiTech \cite{zhang2016single}, UCSD \cite{4587569}, UCF-CC-50 \cite{6619173}, UCF-QNRF \cite{idrees2018composition}, MALL \cite{chen2012feature}, WorldExpo'10 \cite{zhang2015cross}, NWPU \cite{wang2020nwpu},  and JHCrowd \cite{sindagi2020jhu}.
%They contain diverse real scenes and crowd densities and play a pivotal role in crowd-counting research. 
In addition to real-scene datasets, \cite{Wang2019Learning} proposed a large synthetic single-image crowd-counting dataset called GCC. However, GCC is a single-image dataset and is not suitable for multi-view crowd vision tasks. In contrast, SynMVCrowd aims to serve as a challenging platform for both multi-view and single-image crowd counting and localization tasks.

% \textit{In summary, existing multi-view crowd datasets from real scenes face challenges such as a scarcity of annotated images, fixed camera numbers and layouts, low crowd density, and small scene sizes. Synthetic scene datasets have shown improvement in annotated image numbers and scene sizes but still fall short of providing comprehensive diversity. Consequently, this paper proposes a large-scale multi-view synthetic dataset featuring expansive scenes, numerous annotated images, variable camera viewing angles and layout configurations, and a notably high average crowd density. SynMVCrowd aims to serve as a challenging platform for both multi-view and single-image crowd counting and localization.}

\begin{figure}[t]
\begin{center}
%\vspace{-0.4cm}
   \includegraphics[width=\linewidth]{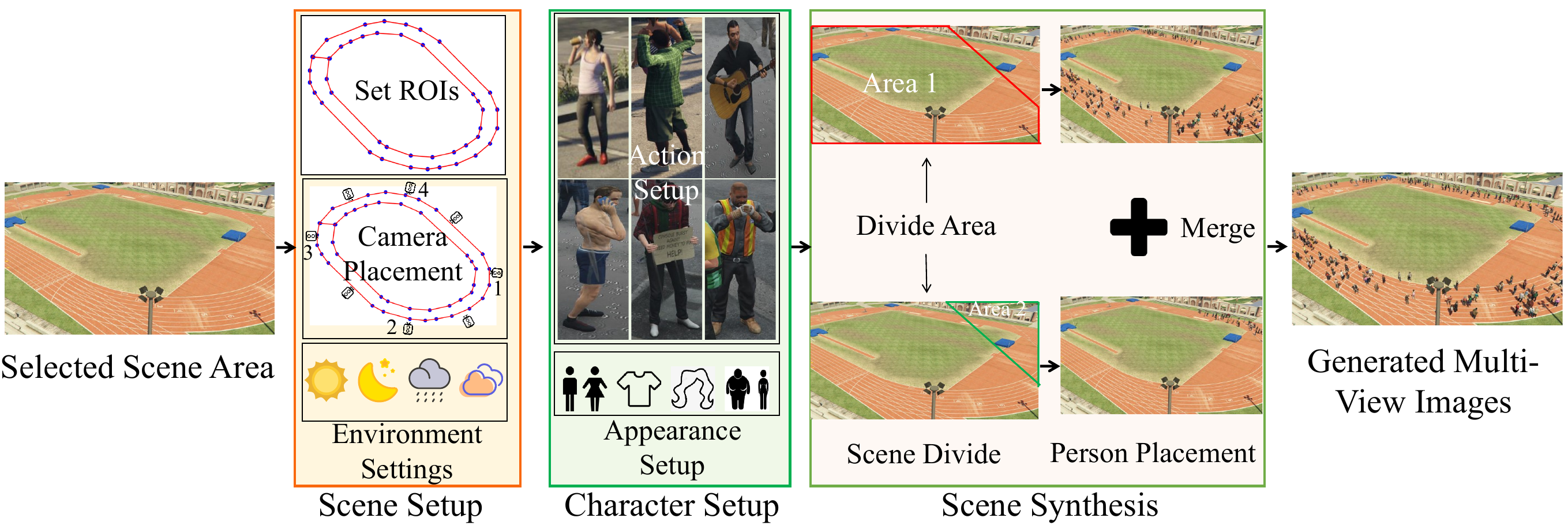}
\end{center}
%\vspace{-0.4cm}
   %\caption{The dataset samples of different scale variations: `Large', `Medium' and `Small'.}
  \caption{The whole process of the multi-view crowd image generation, including scene setup, character setup, and the scene synthesis.}
\vspace{-0.5cm} 
\label{fig: scene generation}
\end{figure}  

\section{Benchmark Creation and Analysis}
%With the purpose of aiding multi-view and single-image crowd tasks in complex environments, we propose a large synthetic crowd dataset called SynMVCrowd. 
%This dataset uses Grand Theft Auto V (GTAV), a modern game engine that simulates a functional city and its surrounding amenities in a photo-realistic three-dimensional world. Benefiting from the excellent simulation performance of the game engine, its scene rendering, texture details, weather conditions and so on are very close to the real world. In addition, the resulting diverse person models and accurate tagging information generated from the game make the game engine a suitable choice for generating a large-scale crowd dataset. 
%SynMVCrowd is created based on Grand Theft Auto V (GTAV), which is a modern game engine that simulates a functional city and its surrounding amenities in a photo-realistic three-dimensional world, with close-to-real-world scene rendering, texture details, weather conditions, and so on.
%, making the game engine a suitable choice for generating a large-scale crowd dataset. 
%We use GCC-CL \cite{Wang2019Learning} as a plug-in to collect crowd data from the game GTAV. Compared with the existing crowd datasets, the proposed dataset SynMVCrowd has the advantages of larger scene variations, more crowd numberes, and more perspectives and images. 
We first introduce how to generate the proposed benchmark SynMVCrowd, and then compare its statistics with existing multi-view and single-image crowd datasets.

% \begin{figure}[t]
% \centering
%     \begin{subfigure}[b]{0.4\linewidth}
%         % \centering
%         \includegraphics[width=\linewidth]{figures/Time Stamp Distribution.pdf}
%         \caption{Time stamp distribution}
%         \captionsetup{font={small}}
        
%         \label{fig:timestamp} 
%     \end{subfigure}
%     \begin{subfigure}[b]{0.42\linewidth}
%         % \centering
%         \includegraphics[width=\linewidth]{figures/Weather Condition Distribution.pdf}
%         \caption{Weather distribution}
%         % \captionsetup{font={small}}
        
%         \label{fig:weather condition}
        
%     \end{subfigure}%

%     \caption{The two pie charts respectively illustrate the distribution of time stamps and weather conditions in SynMVCrowd. In the left one, each label denotes a fixed period of 24 hours a day, such as "0$\sim$3" indicates the time period during (0:00, 3:00).}
%     \label{fig:distribution}
% \end{figure}

% \begin{figure}[t]
% \begin{center}
%    \includegraphics[width=0.8\linewidth]{figures/train_size.pdf}
% \end{center}
% \vspace{-0.2cm}
%    \caption{The scene area of training scenes.}
% \vspace{-0.2cm}
% \label{fig:trainsize}
% \end{figure}

% \begin{figure}[t]
% \begin{center}
%    \includegraphics[width=0.8\linewidth]{figures/other_size.pdf}
% \end{center}
% \vspace{-0.2cm}
%    \caption{The scene area of test and validation scenes.}
% \vspace{-0.2cm}
% \label{fig:testsize}
% \end{figure}

\subsection{SynMVCrowd Benchmark Creation} \label{subsec:SynMVCrowd benchmark creation}
%We collect the frames from GTAV following \cite{Wang2019Learning}. Similarly, the pipeline of data collection consists of three parts: construct scenes, place persons, and scenes synthesis for the congested crowd.  However, different from \cite{Wang2019Learning}, we are interested in capturing information from multiple camera views. This requires to extend the approach by Wang \textit{et al.} \cite{Wang2019Learning}. In this section, we will describe the details of collecting the data in the following.

We extend the GTA-V synthetic platform of GCC \cite{Wang2019Learning} for generating multiple view frames, the full process is illustrated in Fig. \ref{fig: scene generation}, 
consisting of three steps: scene setup, character setup, and scene synthesis.
The scene setup step first defines the Region of Interest (ROI) for placing the people and excludes some invalid regions where people are unlikely to appear, and sets up the camera locations to cover the whole scene as much as possible, and the environment settings, eg, weather and timestamps. In the character setup step, we define the action and appearance of the character in the scene on the GTA-V platform. For scene synthesis, we first divide the whole scene into several areas, then we place the crowd in each area, and record the crowds with the camera views. Finally, we merge different crowd areas to form a large crowd and generate multi-view images of the crowd.

% \red{
\subsubsection{Scene Setup}
In the scene setup step, we first select a possible area as the scene background, and set up ROIs to place the crowd. Then, we place the cameras around the scene ROIs to cover the whole scene as much as possible. At last, we also need to set up the environmental conditions, such as weather, light, or timestamps.
% }

% \red{
\textbf{ROIs setup.}
In each scene, we manually design a Region of Interest (ROI) area for placing the crowd.
% Camera locations of each scene are provided in the dataset.
First, we carefully design the scene shape of each scene to make the scenes contain as many people as possible. 
% and show the corresponding results in Figure \ref{fig:train size} to \ref{fig:other size}. 
At the same time, to prevent unreasonable situations where people are standing on the top of buildings or obstacles, we need to define a suitable area of the scene for the crowd placement. Specifically, when constructing each scene, we mark a number of boundary points in the scene so that they are connected to form a specific area to limit the area generated by the character models, as seen in `Set ROIs' of Fig. \ref{fig: scene generation} Top Left. 
% }
%The crowd can only randomly appear in the ROIs, and regions outside of them are set as background. 

% \red{
% For each scene, we elaborately define the Region of Interest (ROI) for placing the persons and exclude some invalid regions where people are unlikely to appear. Specifically, we use the method of connecting multiple points to form a polygon by moving a game role in the scene. The program automatically connects the points in a sequence set by the user. 
% The ROI regions for each scene where people can be generated are shown in Figure \ref{fig:trainsize} and \ref{fig:testsize}. The crowd can only randomly appear in the ROIs, and regions outside of them are set as background.
% }

% \red{
\textbf{Camera placement.}  
Based on the scene ROIs, we set up the camera locations around the ROIs.
First, we use four camera views to observe the scene in the four directions of east, south, west, and north, as the cameras 1-4 in Fig. \ref{fig: scene generation} Top Left. The angle difference between each adjacent camera view pair is approximately 90 degrees. Based on the first four cameras' orientations and locations, we built a camera sequence to circle around the scene. In the sequence, each camera faces the center of the scene area, forming a circular zone around the scene. The offset angle between each adjacent camera view pair is approximately 10 degrees, and the field of view of the camera is 40 degrees. Overall, in the camera sequence, we create 46 different new camera perspectives, each with similar heights and pitch angles, as well as different locations and rotation angles. Including the first four camera views, there are a total of 50 camera views.

\textbf{Environment settings.}
We construct 50 scenes that frequently occur in real life, such as parks, walking streets, beaches, shopping centers, churches, and so on.
%In order to construct more realistic data, we deliberately selected 50 scenes that frequently occur in real life from the fiction city. For example: parks, walking streets, beaches, shopping centers, churches, and so on.
Besides, we include various weather conditions (clear, cloudy, rainy, thunderstorm, foggy, overcast, and extra sunny) and light variations (0-24h) in the scene to increase the diversity of the background environments.
Fig. \ref{fig:display} illustrates the exemplars under various kinds of weather at different times. The two pie charts in Fig. \ref{fig:distribution} respectively show the proportional distribution of weather conditions and timestamps for the proposed dataset:

\begin{figure}[t]
% \small
\begin{center}
   \includegraphics[width=0.9\linewidth]{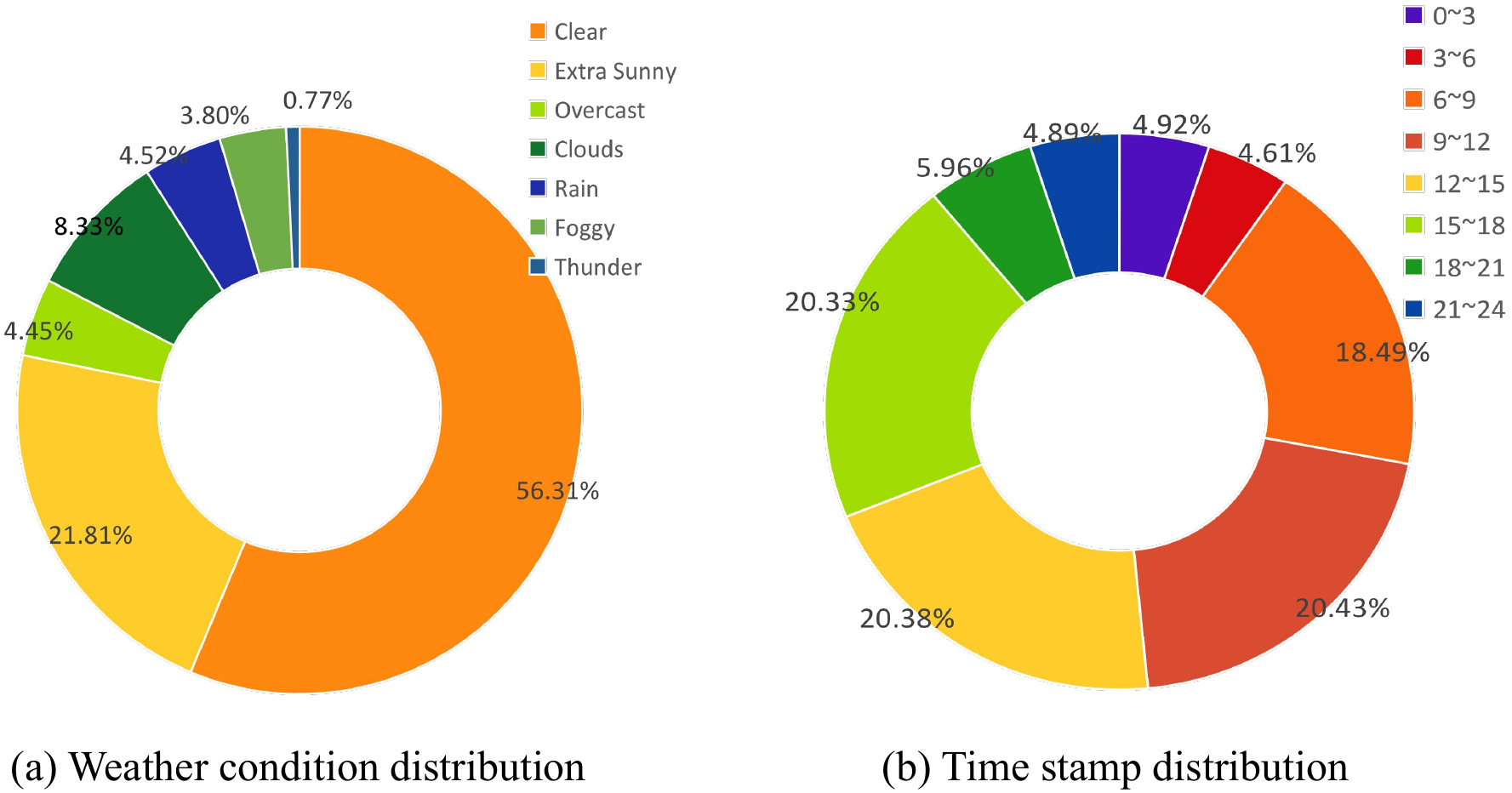}
\end{center}
% \vspace{-0.2cm}
   \caption{The two pie charts respectively illustrate the distribution of time stamps and weather conditions in SynMVCrowd. On the right side, each label denotes a time of 24 hours.}
% \vspace{-0.4cm}
\label{fig:distribution}
\end{figure}

% \red{
\begin{itemize}
    \item \textbf{Weather}: The GTA-V game engine provides 7 different types of weather conditions, including Clear, Extra Sunny, Overcast, Clouds, Rain, Foggy, and Thunder, and 0-24 hours timestamps. We use hook scripts to control all these weather and time conditions as the background environment. Specifically, we firstly set `Clear' as the most common weather conditions, which account for 50\% images in the dataset. Then, using different weather conditions as the control group, we employ a randomization function to assign all weather conditions to the remaining 50\% of the dataset according to a predefined weighting ratio. Based on synthetic crowd image rendering results, the weighting distribution is defined as 2:1:1:1:1:1 for the following conditions: extra sunny, clear, overcast, cloudy, rain, and foggy. And after removing examples with incorrect label information, finally, as shown in Fig. \ref{fig:distribution} (a), the distribution probabilities of all weather are: Clear 56.31\%, Extra Sunny 21.81\%, Overcast 4.45\%,  Clouds 8.33\%, Rain 4.52\%, Foggy 3.80\%, and Thunder 0.77\%. Due to the rendering effect of the crowd under the thunder weather, we only adopt this weather in some special scenarios, accounting for only 0.77\% of all data.
    \item \textbf{Timestamp}: Similarly, the time conditions also use the same strategy to control. Specifically, we divide all time conditions into five main parts: Morning (from 6 a.m. to 9 a.m.), Noon (from 9 a.m. to 12 p.m.), Afternoon (from 12 p.m. to 3 p.m.), Sunset (from 3 p.m. to 6 p.m.), and Evening (from 6 p.m. to 6 a.m.). Different parts share the same weight ratio (defined as 1:1:1:1). %\zqnote{similar questions, the distributions are not the same as the weight ratio. What's their relationship?}
    In addition, Evening time is longer than other time conditions, so we additionally divide time periods, including night (from 6 p.m. to 9 p.m.), nightfall (from 9 p.m. to 12 p.m.), midnight (from 12 p.m. to 3 a.m.), and early morning (from 3 a.m. to 6 a.m.). Among Evening time, different periods also share the same weighting distribution. %\zqnote{similar questions, the distributions are not the same as the weight ratio. What's their relationship?}
    Finally, after filtering data with incorrect labels, as shown in Fig. \ref{fig:distribution} (b), the distribution of timestamps in each part is approximately 20\%. From 6 p.m. to 6 a.m., the average for each timestamp is approximately 5\%.
\end{itemize}
% }
%We choose 50 camera views in each scene to record the crowd information. 

% \zq{Specifically, we use a random function to control all weather conditions to occur randomly in a specific weight ratio. Among them, as shown in Fig. \ref{fig:distribution}, the probability of clear is the highest because the rendering effect of the crowd is the best under such weather conditions.
% On the contrary, the probability of thunderstorms is the lowest because the rendering effect of the crowd is poor. Other weather conditions are randomly generated in an equal probability manner.}
% \zqnote{we need to explain how we decide the distributions of the scene types, whether, and the timestamps.}
% %Further, we alter the weather conditions, and the time of day to increase the diversity of the background environments. According to the effect of the image rendering, we choose seven kinds of weather conditions: clear, cloudy, rainy, thunderstorm, foggy, overcast, and extra sunny. What's more, we randomly select the time period of the scene under different weather conditions. 
% %Due to the situation of thunderstorms and midnight with a poor rendering, we reduced the possibility of choosing the above conditions, while increasing the possibility of choosing clear, extra sunny, and daytime conditions. 
% %During the generation process, we tend to produce more images under common conditions. 

\begin{figure}[t]
% \small
\begin{center}
   \includegraphics[width=0.7\linewidth]{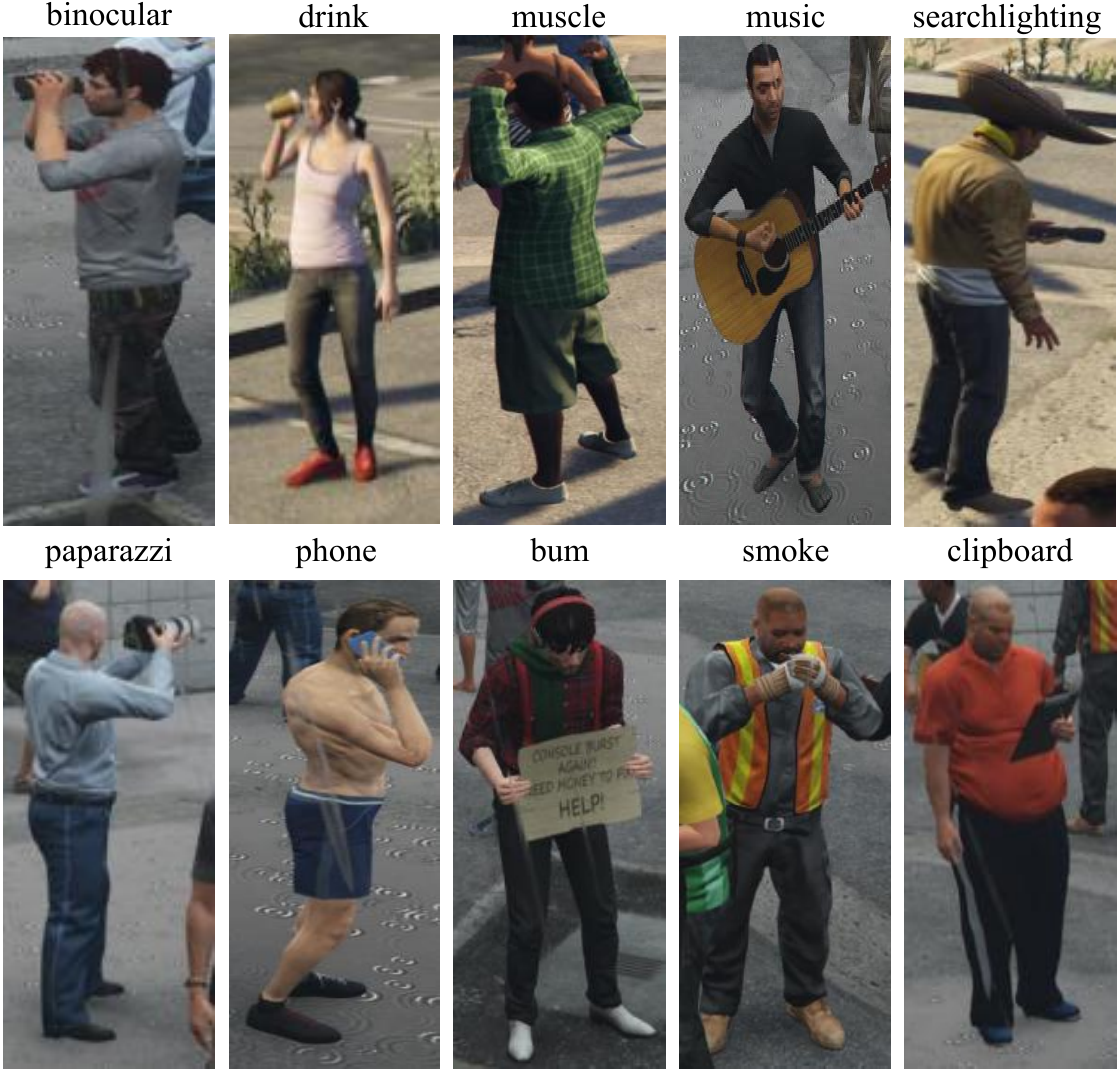}
\end{center}
% \vspace{-0.2cm}
   \caption{The action catalog of SynMVCrowd, consisting of various human poses in daily life.}
% \vspace{-0.4cm}
\label{fig:action}
\end{figure}

\begin{figure}
\begin{center}
   \includegraphics[width=0.8\linewidth]{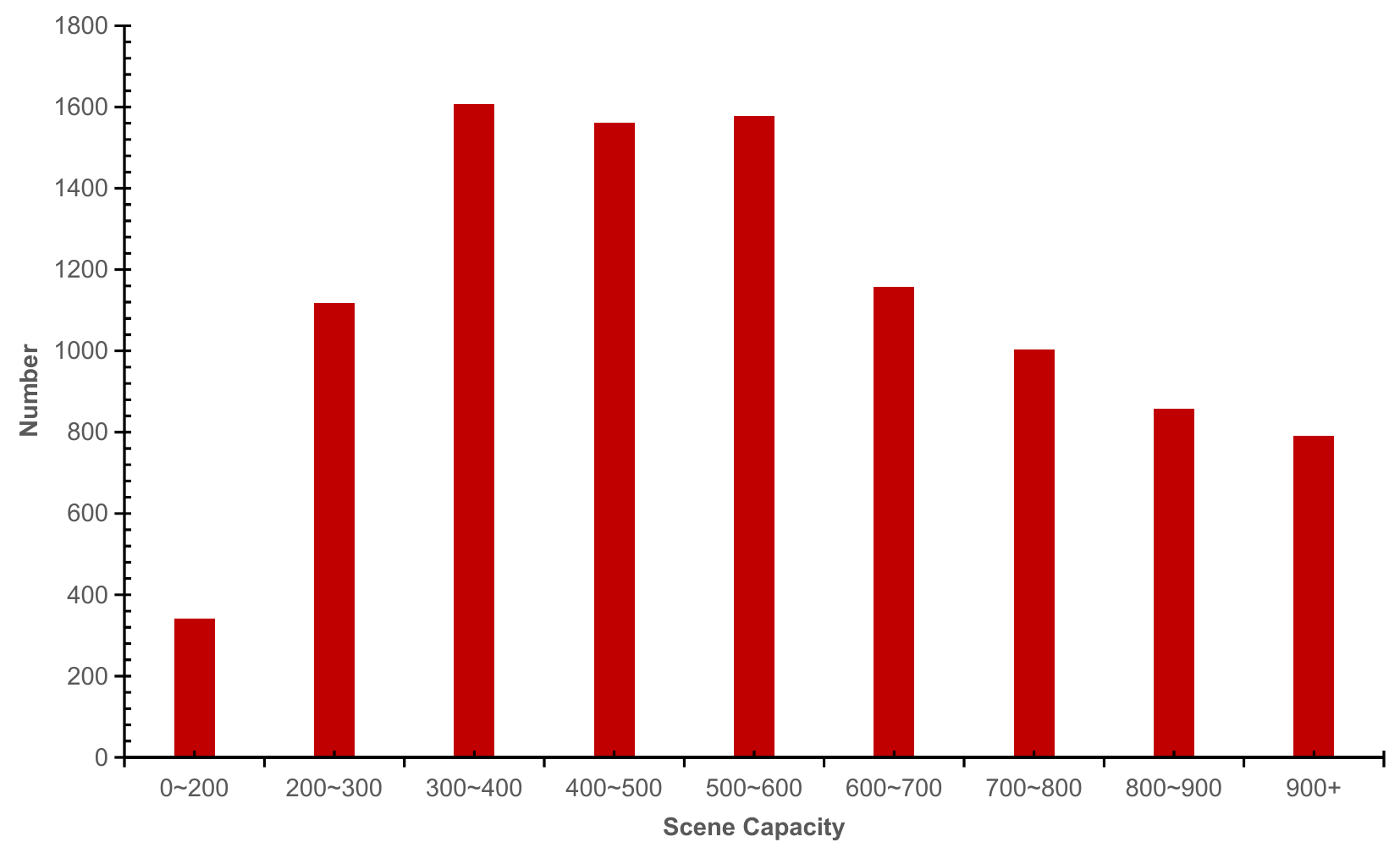}
\end{center}
\vspace{-0.1cm}
   \caption{The frame numbers of crowd count distributions of SynMVCrowd.}
\vspace{-0.2cm}
\label{fig:scene capacity}
\end{figure}

% \textcolor{red}{
\subsubsection{Character Setup} 
% }
% \textcolor{red}{
In GTA-V, it provides 265 character models, each character designed to exhibit distinct physical characteristics that enhance the authenticity of the in-game world. For instance, each model has different physical characteristics, such as skin color, hairstyle, clothing, body shape, gender, \textit{etc}. Additionally, the characters have varied physical builds: ranging from `thin' and `sturdy' to `obesity'. Variations in height and body proportions further ensure that no two characters feel identical. To enable precise tracking and distinguish these characters, each person has a specific ID for mapping their coordinates in the world coordinate system and their locations in each camera-view image. Besides, in order to make the stationary character models more realistic, we give each person a random action. These actions are carefully designed to avoid disrupting the fixed positions of the characters. Unlike dynamic actions (e.g., running, jumping) that involve movement, these actions are subtle and localized. As shown in Fig. \ref{fig:action}, characters adjust the movements of their upper body, such as smoking, drinking, music, and so on. What's more, the distribution of crowd numbers is based on their space capacity to increase the number of people present in each scenario (see Fig. \ref{fig:scene capacity}).
% } 

\subsubsection{Scene Synthesis} 

% \textcolor{red}{
\textbf{Scene divide and person placement.}
After the scene setup and character setup are successful, we will place people in the ROI area. GTA-V has a maximum number limit (256) of people in a scene. To synthesize more congested crowd scenes, we generate the scene step by step. Specifically, as shown in Fig. \ref{fig: scene generation}, we first divide the scene into two areas: Area 1 and Area 2. Next, we will place part of the character models in Area 1 and record this area with 50 cameras immediately. Next, we delete all the character models in that area and place other character models in Area 2, and record the camera views again. If the area contains more character models, the divided area will continue to be split, and the area of each region will further shrink.
% }

% \textcolor{red}{
\textbf{Scene merge.}
Eventually, we integrate all the area images into one scene and produce a set of multi-view frames for the scene. When merging, the area image is a scene image with partial crowd information, and the background image is a scene image that fully covers the crowd regions. The difference between the two images represents the pixel values of the crowd. By recording the pixel differences between all areas and the background image and adding them to the same background image, we can obtain the synthetic scene image. And during the process of picture recording, we adjusted the time flow rate of the game to ensure that the background of the pictures in different areas did not change.

\begin{figure}[t]
\tiny
\begin{center}
   \includegraphics[width=\linewidth]{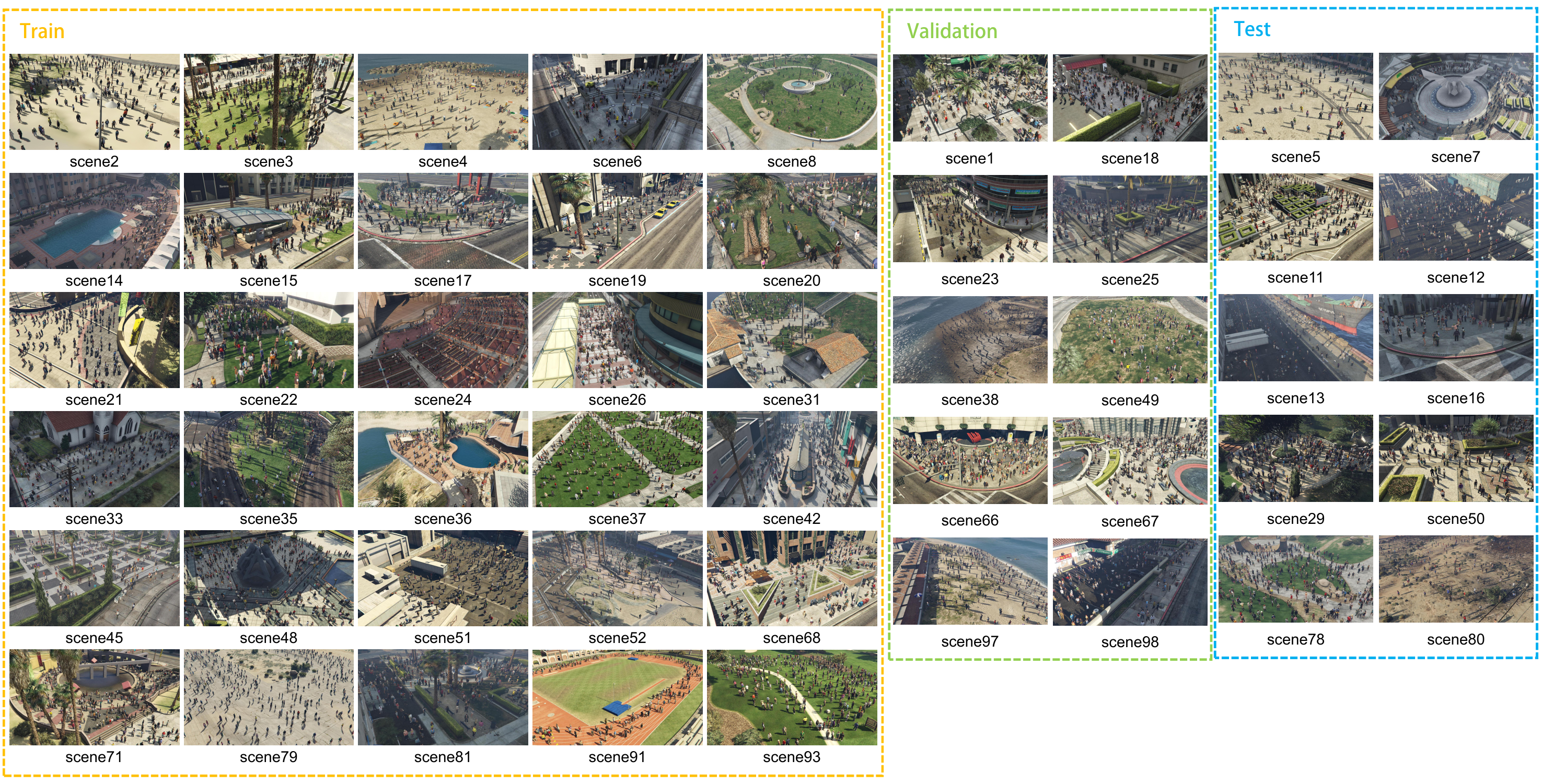}
\end{center}
%\vspace{-0.5cm}
   \caption{The scene examples of SynMVCrowd. The orange box contains a perspective from the training set. The green and blue boxes contain a perspective from the validation and test sets.}
\vspace{-0.4cm}
\label{fig:all scenes}
\end{figure}

% % \begin{figure}[h]

% % \begin{center}
% %    \includegraphics[width=0.8\linewidth,height=0.6\linewidth]{figures/train size.pdf}
% % \end{center}
% % % \vspace{-0.5cm}
% %    \caption{The size of each scene from the training set. The red lines represent the boundary area of the scene, and the blue dots represent each boundary point.}
% % \vspace{-0.4cm}
% % \label{fig:train size}
% % \end{figure}

% % \begin{figure}
% % \small
% % \begin{center}
% %    \includegraphics[width=0.8\linewidth,height=0.6\linewidth]{figures/other size.pdf}
% % \end{center}
% % %\vspace{-0.5cm}
% %    \caption{The size of each scene from the validation set and test set. The red lines represent the boundary area of the scene, and the blue dots represent each boundary point.}
% % \vspace{-0.5cm}
% % \label{fig:other size}
% % \end{figure}

\begin{table*}[t]
\centering
\begin{center}
\scriptsize
\caption{The statistical comparison of the proposed SynMVCrowd and six existing multi-view crowd datasets. 
%\zqnote{add new multi-view datasets}
}
\begin{tabular}{l@{\hspace{0.01cm}}c@{\hspace{0.1cm}}c@{\hspace{-0.15cm}}c@{\hspace{-0.06cm}}c@{\hspace{0.12cm}}c@{\hspace{0.12cm}}c@{\hspace{0.12cm}}c@{\hspace{0.12cm}}c@{\hspace{0.12cm}}c@{\hspace{0.12cm}}c@{\hspace{0.12cm}}}
\toprule
    \multirow{2}{*}{Dataset}  & \multirow{2}{*}{Type} & \multirow{2}{*}{Scenes} &\multirow{2}{*}{Size} & \multirow{2}{*}{Cameras} & \multirow{2}{*}{Frames} & Average  &  \multicolumn{4}{c}{Counting Statistics}\\ \cmidrule{8-11}
%\hline
            &    &     & &    &      & Resolution        
                              & Total        & Min   & Avg      & Max\\
\midrule
    PETS2009 \cite{ferryman2009pets2009}   
    & Real    & 1      & - & 3       & 1,899   & 576$\times$768      & -         & 20   & -     & 40 \\

    DukeMTMC \cite{ristani2016performance}   
    & Real    & 1      & - & 4       & 989  & 1080$\times$1920     & -         & 10   & -     & 30 \\

    Wildtrack \cite{chavdarova2018wildtrack}  
    & Real    & 1      & 12$\times$36 & 7   & 400  & 1080$\times$1920      & -    & -    & 20   & - \\

    MultiviewX \cite{hou2020multiview} 
    & Syn    & 1   & 16$\times$25    & 6       & 400  & 1080$\times$1920      & -         & -   & 40   & - \\

    CityStreet \cite{zhang2019wide} 
    & Real   & 1      & 58$\times$72  & 3       & 500  & 1520$\times$2704      & 64K      & 70   & 128   & 150 \\

    % CVCS \cite{zhang2021CVCS}      
    % & synthetic    & 31  & 10$\times$20 $\sim$ 90$\times$80 & 60-120  & 3,100 & 1080$\times$1920    & 418.5K  & 90   & 135   & 180 \\
    CVCS \cite{zhang2021CVCS}      
    & Syn    & 31  & 90$\times$80 & 60-120  & 3,100 & 1080$\times$1920    & 418.5K  & 90   & 135   & 180 \\
    
    % SynMVCrowd  & synthetic    & 50  & 20$\times$50 $\sim$ 100$\times$120 & 50      & 10,000 & 1080$\times$1920    & 5.3M & 151  & 530  & 1,000 \\    
    SynMVCrowd  & Syn    & 50  & 100$\times$120 & 50      & 10,000 & 1080$\times$1920    & 5.3M & 151  & 530  & 1,000 \\
\botrule
\end{tabular}
\vspace{-0.5cm}
\label{table:Multi-view_datasets}
\end{center}
\end{table*}

\subsection{SynMVCrowd Analysis and Comparison}
We generate a series of scenes that frequently occur in real life to enrich the data distribution of the dataset. As shown in Figure \ref{fig:all scenes}, SynMVCrowd contains 15 different scene types and consists of 50 different urban scenes. Among them, the number of scene types, such as curbside, park, and beach, is larger than other scene types, because these scenarios are widespread in our daily lives. According to different scene types, we divide the training set, test set, and validation set in a ratio of 3:1:1. The training set includes 30 scenes and 14 scene types in total. There are 10 scenes in the validation set and the test set, respectively, where the test set contains more scenario types than the validation set. 
% \red{
Each scene contains 200 multi-view frames, comprising 50 camera views and 10 thousand different images.
In total, SynMVCrowd consists of 500,000 images. And the training set contains 300k images, the test set and validation set contain 100k images (scene number * camera views * frame number).
% }
We show the visualization result of the divided dataset in Figure \ref{fig:all scenes}.

\begin{figure}[t]
\begin{center}
    \includegraphics[width=0.8\linewidth]{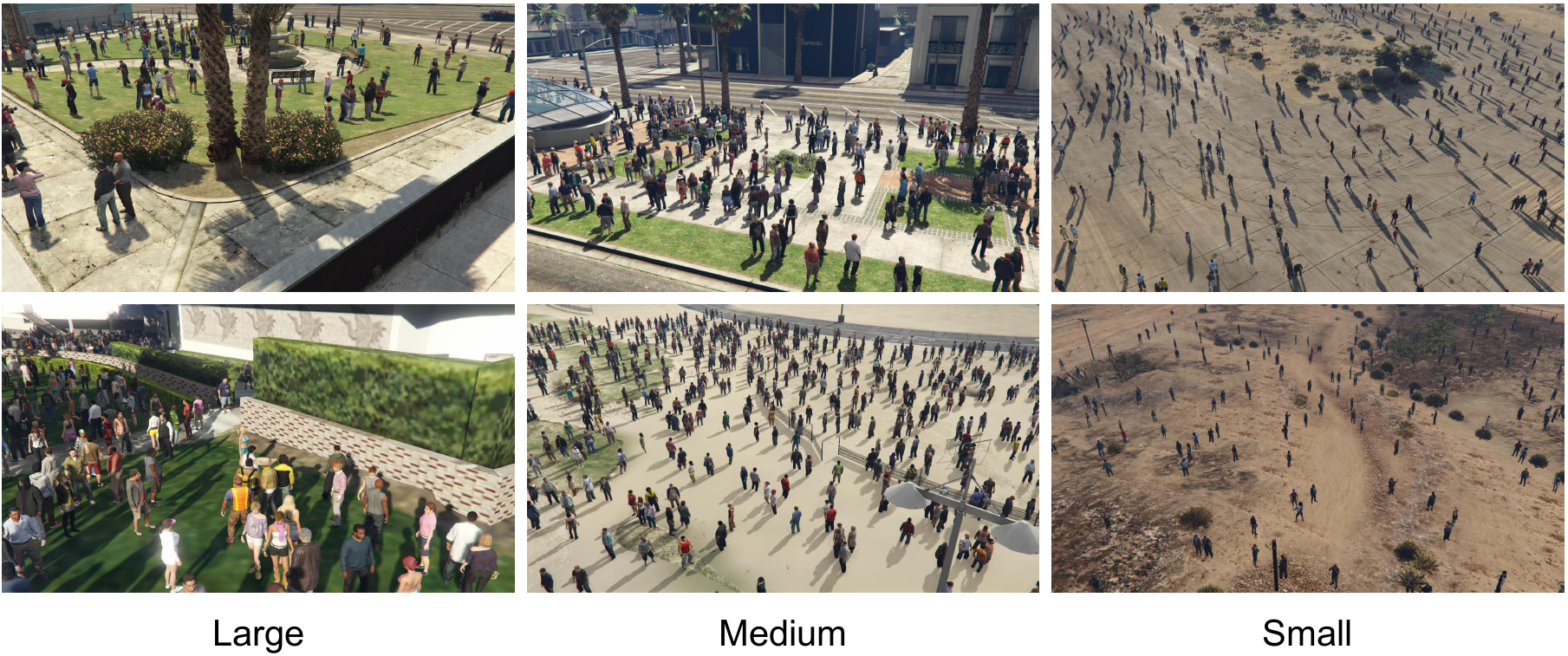}
\end{center}
% \vspace{-0.5cm}
   %\caption{The dataset samples of different scale variations: `Large', `Medium' and `Small'.}
  \caption{The dataset samples of scale variations.}
\vspace{-0.3cm}
\label{fig:ACM-rebuttal-1}
\end{figure}

What's more, we carefully considered the scale variation in the real world, since we contain scenes of variable sizes, and the crowds in the scenes vary with scale, as in Figure \ref{fig:ACM-rebuttal-1}. 
%The corresponding results are shown in Figure \ref{fig:ACM-rebuttal-1}. 
We also generate accurate crowd semantic segmentation images and crowd coordinates in the image and the world coordinate system, with a special ID for each person. Furthermore, we record all auxiliary information in each scene in detail, which includes camera parameters, weather conditions, time, scene size \textit{etc}.
For each scene, we set up 50 camera views to record the crowd information of the scene and repeat 200 times to capture different crowd distributions under various weather conditions in the scene. Thus, each scene contains 200 multi-view frames, comprising 50 camera views. In total, the whole dataset consists of 500,000 images, with a resolution of $1920 {\times} 1080$ pixels, each image contains crowds ranging from 200 to 1000, with an average of more than 500 people.

\begin{table}[t]
\centering
\small
\caption{The statistic comparison of the proposed SynMVCrowd and nine existing single-image crowd datasets.}
\vspace{-0.2cm}
\setlength\tabcolsep{4pt}
\begin{tabular}{lcccccccc}
\toprule
\multirow{2}{*}{Dataset} & \multirow{2}{*}{Type} & \multirow{2}{*}{Year} & Average & Number & \multicolumn{4}{c}{Counting Statistics}  \\ 
\cmidrule{6-9}
& & & Resolution & of Images & Total & Min & Avg & Max  \\  
\midrule
USCD \cite{4587569}       & Real & 2008 & 158$\times$238      & 2,000            & 50K       & 11          & 25             & 46 \\
Mall \cite{chen2012feature}       & Real & 2012 & 240$\times$320      & 2,000            & 62K       & 13          & 31             & 53 \\
UCF\_CC\_50 \cite{6619173} & Real & 2013 & 2101$\times$2888    & 50               & 64K       & 94          & 1,280          & 4,543 \\
WorldExpo'10 \cite{zhang2015cross} & Real & 2015 & 576$\times$720      & 3,980            & 200K      & 1           & 50             & 253 \\
ShanghaiTech Part A \cite{zhang2016single} & Real & 2016 & 589$\times$868    & 482         & 242K      & 33          & 501            & 3,139  \\
ShanghaiTech Part B \cite{zhang2016single} & Real & 2016 & 768$\times$1024   & 716         & 88K      & 9           & 27             & 57     \\
GCC \cite{Wang2019Learning}         & Syn & 2019 & 1080$\times$1920    & 15,212      & 7.6M    & 0           & 501            & 3,995  \\
UCF\_QNRF \cite{idrees2018composition}   & Real & 2019 & 2013$\times$2902    & 1,535            & 1.3M    & 49          & 815            & 12,865 \\
NWPU-Crowd \cite{wang2020nwpu}  & Real & 2020 & 2191$\times$3209    & 5,109            & 2.1M    & 0           & 418            & 20,033 \\
SynMVCrowd   & Syn & 2024 & 1080$\times$1920    & 500,000          & 5.3M  & 47         & 530            & 1,000 \\
\bottomrule
\end{tabular}
\vspace{-0.6cm}
\label{table:Single-image_datasets}
\end{table}

\textbf{Comparison with multi-view datasets.}
In order to further demonstrate the superiority of our proposed dataset, we compare the proposed SynMVCrowd dataset with other multi-view crowd datasets in Table \ref{table:Multi-view_datasets}. PETS2009 \cite{ferryman2009pets2009}, DukeMTMC \cite{ristani2016performance}, Wildtrack \cite{chavdarova2018wildtrack} and CityStreet \cite{zhang2019wide} are all single-scene real-world datasets. Due to the difficulty of collecting realistic data, the distribution of crowd numbers and background environments in these four realistic datasets is simple. PETS2009 \cite{ferryman2009pets2009} and DukeMTMC \cite{ristani2016MTMC} contain a small number of crowd ($<$40). In Wildtrack \cite{chavdarova2018wildtrack} and MultiviewX \cite{hou2020multiview}, the scale of the scene is relatively small, only within 12$\times$36 meters area. In CityStreet \cite{zhang2019wide}, though the scene size and crowd numbers have been enlarged, a scene with only 3 camera views is relatively small. CVCS \cite{zhang2021CVCS} is a large synthetic dataset that can provide data on the distribution of different populations in multiple scenarios under different background environments, but each scene contains a sparse crowd with small crowd numbers. 
%Notably, CVCS \cite{zhang2021CVCS} also collected data from GTAV. But compared to their work, since the number of people in CVCS range from 90 to 180, they don't have to worry about the problem of the maximum number limit. As a result, they can place character models directly in the scene and capture still frames without the need for scene synthesis operations (see Sec \ref{subsec:SynMVCrowd benchmark creation} for details), which will greatly reduce the difficulty of dataset collection. 
In contrast, the proposed SynMVCrowd contains more scenes, more crowds, higher densities, and more frames, which are more suitable for comprehensively validating crowd vision tasks.

% \begin{figure}[t]
% \small
% \begin{center}
%    \includegraphics[width=0.9\linewidth]{figures/annotation.pdf}
% \end{center}
% % \vspace{-0.2cm}
%    \caption{The data annotation of SynMVCrowd: view images (first row) and semantic segmentation (second row).}
% % \vspace{-0.4cm}
% \label{fig:annotation}
% \end{figure}

\textbf{Comparison with single-image datasets.}
Compared with the existing single-image crowd-counting datasets, SynMVCrowd is the largest one from the perspective of the image level. Table \ref{table:Single-image_datasets} summarizes the statistics of the single-image crowd-counting datasets. Before 2016, many datasets had the disadvantage of low resolution, such as USCD \cite{4587569}, Mall \cite{chen2012feature}, and WorldExpo'10 \cite{zhang2015cross}, whose average resolution was less than 600$\times$800, and the number of images was less than 4,000. 
% usually show worse visual quality, which may affect the model to achieve higher count accuracy. Besides, such as UCF\_CC\_50 \cite{6619173}, 
On the other hand, UCF\_CC\_50 \cite{6619173}, ShanghaiTech Part A and B \cite{zhang2016single} with less than 1,000 images will easily cause over-fitting problems. 
In more recent datasets, GCC \cite{Wang2019Learning}, UCF\_QNRF \cite{idrees2018composition}, and NWPU-dataset \cite{wang2020nwpu} have a wider range of people and higher resolution. Compared with previous works, GCC is a synthetic dataset with the largest number of images, around 15,000 images. However, compared to previous single-image datasets, SynMVCrowd can provide 5.3 million images for training, with various scene and perspective variations, weather, and lighting changes. 
%Notably, SynMVCrowd contains 50 multi-views for each scene and can be used for multi-view crowd vision tasks.
Overall, the proposed SynMVCrowd benchmark is a challenging platform for existing multi-view and single-image crowd counting and localization methods, which shall advance the research of crowd counting and localization to more practical applications.

\begin{figure*}[t]
\begin{center}
   \includegraphics[width=\linewidth]{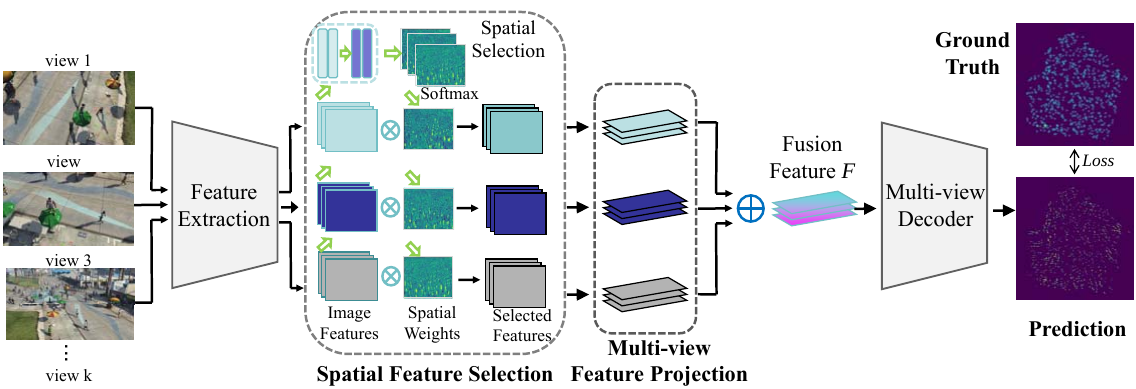}
\end{center}
%\vspace{-0.1cm}
   \caption{The pipeline of multi-view localization and counting baseline model: single-image feature extraction, spatial feature selection, multi-view feature projection and fusion, and multi-view decoding with MSE or Optimal Transport (OT) loss.}
   %\NOTE{(Redraw the figure: flat)}
\vspace{-0.4cm}
\label{fig:supp_pipeline}
\end{figure*}

\section{SynMVCrowd Benchmark Evaluation}
In this section, we evaluate and compare the SOTA multi-view and single-image crowd counting and localization methods on the proposed SynMVCrowd benchmark. Besides, we propose a strong multi-view crowd counting and localization baseline, outperforming all existing methods.

% \begin{table*}[t]
% % \small
% \centering
% \begin{tabular}%{c|ccc|c@{\hspace{0.15cm}}c@{\hspace{0.15cm}}c@{\hspace{0.15cm}}c@{\hspace{0.15cm}}c@{\hspace{0.10cm}}}
% {c|ccc|ccccc}
% \hline
%     Method    & MAE $\downarrow$ & NAE $\downarrow$ & MSE $\downarrow$ & MODA $\uparrow$   & MODP $\uparrow$       & Precision $\uparrow$       & Recall $\uparrow$    & F1\_score  $\uparrow$   \\
% \hline
%     CVCS backbone \cite{zhang2021CVCS}   & 117.02 & 0.246  & 157.04 & - & - & - & - & - \\
%     CVCS \cite{zhang2021CVCS}       & \textbf{82.45}  & \textbf{0.173}  & \textbf{108.61}  & - & - & - & - & - \\
% \hline
%     MVDet \cite{hou2020multiview}       &   - & - & - &     27.0   &  52.2      &  72.2          & 43.9        & 54.60     \\
%     SHOT  \cite{song2021stacked}        &   - & - & - &      32.5   &  52.6      &  74.5          & \textbf{49.3}        & \textbf{59.34}   \\
%     MVDeTr \cite{hou2021multiview}      &   - & - & - &      \textbf{35.6}   & \textbf{ 69.7}      &  \textbf{95.4}          & 37.4        & 53.73   \\
%     3DROM \cite{qiu20223d}       &   - & - & - & 24.2     & 59.2    & 86.1    & 28.8  & 43.16  \\
%     Baseline (Ours)                &   - & - & -   &   - & - & -   &   - & - \\
% \hline
% \end{tabular}
% \caption{The results of multi-view crowd counting and localization methods on the SynMVCrowd dataset. For CVCS and CVCS backbone, we only report the multi-view crowd counting performance. For the rest methods, we only report the multi-view crowd localization performance.}
% \vspace{-0.4cm}
% \label{table:multi_view_results1}
% \end{table*}

\subsection{Multi-view Crowd Localization} \label{subsec:multi-view localization}
%Multi-view crowd localization focuses on predicting crowd localization on the ground of the scenes via synchronized and calibrated multi-cameras.

\subsubsection{Baseline Method} 
%\NOTE{(Add a new paragraph for explaining the baseline method.)}
We propose a strong multi-view localization baseline, whose pipeline is shown in Fig. \ref{fig:supp_pipeline}, which consists of 4 modules: single-view feature extraction, spatial feature selection, multi-view feature projection and fusion, and multi-view decoding.
The single-view feature extraction module uses feature extraction backbone nets (eg, ResNet-18, VGG19, or Transformer) for extracting the features of input images. 
% \red{
The effect of the spatial feature selection module is to select effective features in a spatial-attention fashion. 
Specifically, we input each single-view feature into the spatial feature selection module, and it first estimates the spatial attention map of each view, which is then normalized with a softmax layer across all views. 
The spatial attention map is multiplied by the input view features to select useful features. 
% }
%In the spatial feature selection module, each view's feature is fed into the shared spatial feature selection module to predict a weight map, where each pixel of the weight map indicates the selection likelihood scores. The weight maps for all views act as a selective mask to filter out useless feature information from each view by using a multiplication operation.

Then, the filtered single-view features are projected to the scene ground plane via a projection layer based on the spatial transformation network \cite{Jaderberg2015Spatial}, 
% \red{
where Translation between 3D locations (x, y, z) and 2D image pixel coordinates $(u, v)$ is done via 
$$ s \begin{pmatrix}u\\v\\1\end{pmatrix} = P \begin{pmatrix}x\\y\\z\\1\end{pmatrix} = A \left [ R|t \right ] \begin{pmatrix}x\\y\\z\\1\end{pmatrix}.$$
$s$ is a real-valued scaling factor, and $P$ is a $3$ by $4$ perspective transformation matrix. Specifically, $A$ is the $3$ by $3$ intrinsic parameter matrix. $[R|t]$ is the $3$ by $4$ joint rotation-translation matrix, or extrinsic parameter matrix, where $R$ specifies the rotation and t specifies the translation. The people's height $z=1750mm$ is the same as in MVMS \cite{zhang2019wide}.
% }
The projected features are further fused via a view max-pooling step. After that, a reasonable selection of feature information from different perspectives for fusion can effectively help model prediction.
Finally, the fused multi-view features are fed into the multi-view decoder for predicting the people's occupancy map on the ground. 

As to the baseline method's training loss, we adopt the \textbf{optimal transport} loss, which utilizes point maps as supervision in the single-view localization tasks \cite{Wan2021CVPR} for multi-view crowd localization. %\ky{Optimal transport (OT) is a metric that measures the similarity of two distributions. In the context of crowd localization, the OT loss supervises the network by encouraging the predicted crowd distribution (density map) to closely match the ground-truth crowd locations.}
% \red{
We denote $x$ as the predicted density map, where $x_i$ is the density of pixels. And, we denote $y$ as the ground-truth dot map, where $y_j \in \begin{matrix} 0, 1 \end{matrix}$ indicates there is a person at location $y_j$.
The optimal transport loss is as follows.
$$\mathcal{L}^{\mathcal{T}}_C = \min_{\bf{P} \in {\cal R}_+^{n\times m}} \langle \textbf{C}, \textbf{P} \rangle - \varepsilon H(\textbf{P}) + \tau \parallel P1_{m} - a\parallel_{2}^{2} + \tau \parallel P^{T}1_{n} - b\parallel_{1},$$
where $P\in{\mathcal{R}^{n\times m}_{+}}$ is the transport plan matrix, which indicates each density $x_i$ of predicted density map to $y_{i}$ of the ground-truth dot map for measuring the cost. $C\in{\mathcal{R}^{n\times m}_{+}}$ is the transport cost matrix, where $C_{ij}$ measures the cost of moving the predicted density at $x_i$ to $y_j$. We use Euclidean distance between points as a cost function:
$C_{ij} = c(x_i,y_j) = exp(\parallel x_i - y_j \parallel)$.
The second term $H(\textbf{P})=-\sum_{ij}\textbf{P}_{ij}\log \textbf{P}_{ij} $ if the entropic regularization term. The remaining term is ensuring all predicted density values have a corresponding annotation, and all ground-truth points are used in the transport plan, respectively. Finally, $\tau$ and $\varepsilon$ are weighted factors to adjust the weight of the last three terms.
% }

\subsubsection{Experiment Settings}

\textbf{Implementation details.}
The synthetic dataset contains 50 scenes in total, 30 scenes of which are used for training, 10 for validation, and the remaining 10 for testing. We randomly select 5 camera views 5 times and randomly choose 10 frames out of 200 of a scene in each iteration. For evaluation, we randomly select 5 views for 5 times per frame of each scene. The experimental settings remain consistent with all methods.
% \textcolor{red}{We downsample the input image resolution to 360x640, which is a resize of 1/3 of the original resolution. However, in MVDet, SHOT, MVDeTr, and 3DROM, we always keep the original resolution as input. It is important to note this difference in input resolution to ensure accurate results while working with these methods.}
%In CVCS and CVCS backbone, we downsample the input image resolution to $360\times640$ (resize of 1/3 of the original resolution), but in MVDet, SHOT, MVDeTr, and 3DROM, we keep the original resolution as input.%
We use a patch training method as in \cite{zhang2021CVCS}, with a resolution of 200$\times$200, and each example contains 3 patches. Each grid in the ground plane map stands for 0.2 meters in the real world. During training, we first train the 2D detection task as the pretraining for the feature extraction subnet, where only head annotations in 2D images are used in training. Then, the multi-view decoding subnet and feature extraction subnet are trained together, where the loss term weight follows \cite{hou2020multiview}. In the end, we use the whole image to evaluate the model's performance. Two Nvidia RTX3090 GPUs are used to conduct all experiments. 
In this work, the proposed model is trained with MSE loss or OT loss, respectively. We trained the model for approximately 200 epochs, using Stochastic Gradient Descent (SGD) as the optimizer. Our learning rate was set to a maximum value of 0.0001, and we employed the PyTorch scheduler ``LambdaLR''. We also used a momentum of 0.9 and a weight decay of 1e-4.

\textbf{Comparisons.}
We compare the mean square error loss (MSE) and the optimal transport loss on the baseline model, denoted as `Baseline (MSE)' and `Baseline (OT)'. In addition, the proposed baseline method is also compared with seven state-of-the-art multi-view crowd localization methods, including MVDet \cite{hou2020multiview}, SHOT \cite{song2021stacked}, MVDeTr \cite{hou2021multiview}, 3DROM \cite{qiu20223d}, SVCW \cite{zhang2024multi}, MVOT \cite{zhang2024mahalanobis}, and TrackTacular \cite{Teepe_2024_CVPR} on the proposed SynMVCrowd benchmark.

\textbf{Evaluation metrics.} We use 5 metrics to evaluate the localization precision: Multiple Object Detection Precision (MODP), Multiple Object Detection Accuracy (MODA), Precision, Recall, and F1\_score. First, we count the number of True Positives (TP), False Positives (FP), and False Negatives (FN) to compute the metrics. The formulas of metric can be formulated as follows: $ MODA=1-(FP+FN)/(TP+FN) $, $ MODP=(\sum(1-d[d<t]/t))/TP $,  $ Precision=TP/(TP+FP) $, $ Recall=TP/(TP+FN) $ and $ F1\_score=2Precision*Recall/(Precision+Recall) $. 
%In the above formulas, TP represents the number of Maximum Bipartite Matching between prediction results and ground truth, where the matching condition must be sure that the minimum distance between one predicted point and the ground truth is less than the pre-defined distance threshold. FP and FN respectively indicate false positives and false negatives in the result of Maximum Bipartite Matching. 
F1\_score is a balance of \textit{Precision} and \textit{Recall}. In MODP, $d$ represents the Euclidean distance between each predicted point and the GT point, and $t$ represents the distance threshold.

\begin{figure*}[t]
\begin{center}
   \includegraphics[width=\linewidth]{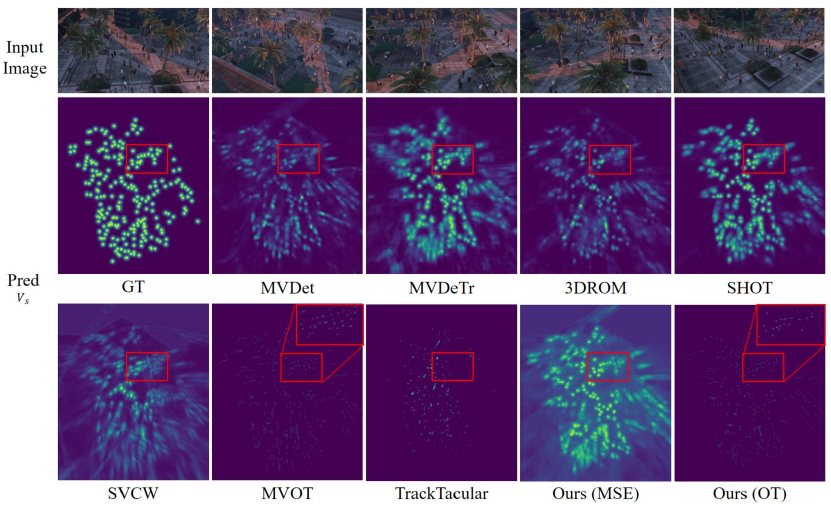}
\end{center}
% \vspace{-0.3cm}
   \caption{The visualization results of multi-view crowd localization: the first row is the camera view input, and the second row is the corresponding ground truth (GT) and prediction results.
      %\NOTE{(Draw a figure of multi-view counting results.)}
   }
\vspace{-0.4cm}
\label{fig:multiveiw_results}
\end{figure*}

\begin{table}[t]
% \small
\centering
\caption{The results of multi-view crowd localization methods on the SynMVCrowd dataset. Our baseline method, Baseline (OT) outperforms all comparisons.}
% \vspace{-0.3cm}
\begin{tabular}%{c|ccc|c@{\hspace{0.15cm}}c@{\hspace{0.15cm}}c@{\hspace{0.15cm}}c@{\hspace{0.15cm}}c@{\hspace{0.10cm}}}
{@{\hspace{0cm}}l@{\hspace{0.08cm}}c@{\hspace{0.08cm}}c@{\hspace{0.08cm}}c@{\hspace{0.08cm}}c@{\hspace{0.08cm}}c@{\hspace{0cm}}}
\toprule
    Method    & MODA $\uparrow$   & MODP $\uparrow$  & Precision$\uparrow$       & Recall$\uparrow$    & F1\_score$\uparrow$  \\
\midrule
    MVDet \cite{hou2020multiview}      &     27.0   &  52.2      &  72.2          & 43.9        & 54.6     \\
    SHOT  \cite{song2021stacked}        &      32.5   &  52.6      &  74.5          & 49.3        & 59.3   \\
    MVDeTr \cite{hou2021multiview}       &      35.6   & 69.7      &  \textbf{95.4}          & 37.4        & 53.7   \\
    3DROM \cite{qiu20223d}        & 24.2     & 59.2    & 86.1    & 28.8  & 43.2  \\
    SVCW \cite{zhang2024multi} &35.8 &55.6 &75.8 & 51.7 & 61.4\\ 
    MVOT \cite{zhang2024mahalanobis} &45.5 &66.3 &83.4 & 56.9 & 67.6\\ 
    TrackTacular \cite{Teepe_2024_CVPR}  &45.8 &71.1 &92.6 & 49.8 & 64.8\\
\midrule
    % Baseline (MSE) &   32.4 & 68.0 & 91.6   &   35.6 & 51.27 \\
    Baseline (MSE) & 34.6 &74.5 &92.9 &37.4 &53.4\\
    Baseline (OT)  &  \textbf{49.6} & \textbf{70.2} & 88.6   &  \textbf{57.0} & \textbf{69.4} \\       
    %&   \textbf{43.8} & 69.6 & 88.4   &  \textbf{51.6} & \textbf{65.03} 
    %\\
\botrule
\end{tabular}
\vspace{-0.5cm}
\label{table:multi_view_localization_results}
\end{table}

\begin{table*}[t]
\scriptsize
\centering
\caption{
% \red{
Comparison of the multi-view people detection performance on Wildtrack and MultiviewX using 5 metrics.
All comparison methods train and test on Wildtrack or MultiviewX (single scene), while ours are trained on SynMVCrowd and finetuned on Wildtrack or MultiviewX or with the domain adaptation technique.
% }
%\ZQNOTE{(Update later.)}
}
\begin{tabular}{@{\hspace{0cm}}l@{\hspace{0.12cm}}|c@{\hspace{0.12cm}}c@{\hspace{0.12cm}}c@{\hspace{0.12cm}}c@{\hspace{0.12cm}}c@{\hspace{0.12cm}}|c@{\hspace{0.12cm}}
c@{\hspace{0.12cm}}c@{\hspace{0.12cm}}c@{\hspace{0.12cm}}c@{\hspace{0cm}}}
\hline
    Dataset &  \multicolumn{5}{c|}{Wildtrack}  &  \multicolumn{5}{c}{MultiviewX}  \\
    Method         & MA.    & MP.        & P.        & R.     & F1.
                   & MA.    & MP.        & P.        & R.     & F1. \\
\hline
    RCNN \cite{xu2016multi}                    & 11.3    & 18.4        & 68         &  43  & 52.7
                                                   & 18.7    & 46.4        & 63.5         &  43.9  & 51.9 \\
    POM-CNN \cite{fleuret2007multicamera}      &  23.2  &    30.5      & 75     & 55  & 63.5
                                                   &  -  &    -      & -     & -  & -       \\
    DeepMCD  \cite{chavdarova2017deep}          & 67.8   & 64.2         & 85         & 82  & 83.5
                                                   & 70.0   & 73.0         & 85.7         & 83.3  & 84.5\\
    DeepOcc. \cite{baque2017deep}              & 74.1    & 53.8        & 95     & 80  & 86.9
                                                   & 75.2    & 54.7        & 97.8     & 80.2  & 88.1\\
    Volumetric \cite{iskakov2019learnable}     & 88.6    & 73.8        & 95.3     & 93.2  & 94.2
                                                   & 84.2    & 80.3        & 97.5     & 86.4  & 91.6\\
%\hline
    MVDet   \cite{hou2020multiview}              & 88.2    & 75.7        & 94.7        & 93.6  & 94.1
                                                   & 83.9    & 79.6        & 96.8        & 86.7  & 91.5\\
    SHOT    \cite{song2021stacked}                & 90.2    & 76.5        & 96.1        & 94.0  & 95.0
                                                   & 88.3    & 82.0        & 96.6        & 91.5  & 94.0 \\
    MVDeTr  \cite{hou2021multiview}            & 91.5    & 82.1        & 97.4        & 94.0  & 95.7
                                                    & 93.7    & 91.3        & 99.5        & 94.2  & 97.8 \\
    3DROM   \cite{qiu20223d}                     & 93.5    & 75.9        & 97.2        & 96.2  & 96.7
                                                     & 95.0    & 84.9        & 99.0        & 96.1  & 97.5 \\

    MVOT \cite{zhang2024mahalanobis}               & 92.1	& 81.3    & 94.5     & 97.8     & 96.1
                                                & 96.7	& 86.1     & 98.8     & 97.9      & 98.3        \\

    EarlyBird \cite{Teepe_2024_WACV}     & 91.2 & 81.8    & 94.9        & 96.3       & 95.6
                                            & 94.2    & 90.1     & 98.6      & 95.7   & 97.1 \\

    TrackTacular \cite{Teepe_2024_CVPR}     & 92.1    & 76.2       & 97.0        & 95.1  &   96.0
                                                & 96.5    & 75.0        & 99.4       & 97.1  &98.2 \\

    POV-MVD \cite{Alturki_2025_CVPR}       &93.6 &82.4 &96.6 &97.0 &96.8
                                          &97.3 &95.0 &99.5 &97.9 &98.7 \\
\hline
\hline
    % Ours (ft)    & 73.9	& 72.4          & 86.8       & 87.2   & 87.0      & 81.1	& 77.2         & 95.0        & 85.6 	& 90.1 \\
    % Ours (ft+da)  & \textbf{78.9}	& 73.6  & 88.7  & 90.4 & \textbf{89.5}  & \textbf{83.8}	& 76.5         & 97.1        & 86.4   & \textbf{91.4} \\
    % Baseline (test)        & 31.93 & 48.46 & 78.36 & 44.12 & 56.45 & 25.40 & 51.96 & 62.588 & 62.00 & 62.44 \\
    % Baseline (10\%)        & 55.04 & 68.27 & 77.64 & 77.31 & 77.47 & 66.35 & 73.04 & 92.60 & 72.12 &81.06 \\
    % Baseline (30\%)        & 67.75 &71.09  & 83.21 & 84.87 & 84.04 & 74.00&75.63  & 92.53 & 80.50 & 86.09\\
    % Baseline (100\%)       & 85.92 &76.53 &94.07 &91.70 & 92.87 & 84.78 &84.66 &99.07 &85.59 &91.84 \\ 
    % Baseline (10\%+DA)     & 63.15 & 73.58 & 82.43 & 83.92 & 83.17  & 72.89 & 78.24 & 94.12 & 78.53 & 85.67 \\
    % Baseline (30\%+DA)     & 75.28 & 78.36 & 89.74 & 89.15 & 89.44 & 79.80 & 81.25 & 95.87 & 86.15 & 90.73 \\
    % Baseline (100\%+DA)    & \textbf{88.73} & \textbf{81.25} & \textbf{95.32} & \textbf{93.85} & \textbf{94.58} & \textbf{87.62} & \textbf{87.89} & \textbf{99.35} & \textbf{88.97} & \textbf{93.88} \\
        Baseline (test)        & 31.9 & 48.5 & 78.4 & 44.1 & 56.5 & 25.4 & 52.0 & 62.6 & 62.0 & 62.4 \\
    Baseline (10\%)        & 55.0 & 68.3 & 77.6 & 77.3 & 77.5 & 66.4 & 73.0 & 92.6 & 72.1 & 81.1 \\
    Baseline (30\%)        & 67.8 & 71.1 & 83.2 & 84.9 & 84.0 & 74.0 & 75.6 & 92.5 & 80.5 & 86.1 \\
    Baseline (100\%)       & 85.9 & 76.5 & 94.1 & 91.7 & 92.9 & 84.8 & 84.7 & 99.1 & 85.6 & 91.8 \\ 
    Baseline (10\%+DA)     & 63.2 & 73.6 & 82.4 & 83.9 & 83.2 & 72.9 & 78.2 & 94.1 & 78.5 & 85.7 \\
    Baseline (30\%+DA)     & 75.3 & 78.4 & 89.7 & 89.2 & 89.4 & 79.8 & 81.2 & 95.9 & 86.2 & 90.7 \\
    Baseline (100\%+DA)    & \textbf{88.7} & \textbf{81.3} & \textbf{95.3} & \textbf{93.9} & \textbf{94.6} & 
    \textbf{87.6} & \textbf{87.9} & \textbf{99.4} & \textbf{89.0} & \textbf{93.9} \\

\hline
\end{tabular}
%\vspace{-0.1cm}
%\vspace{-0.4cm}
\label{table:Wildtrack_MultiviewX_results}
\end{table*}

\subsubsection{Experiment Results}

\textbf{Result Analysis}
The results of SOTA methods and the proposed baselines on SynMVCrowd are presented in Table \ref{table:multi_view_localization_results}, which shows all SOTAs perform badly with lower MODA and F1\_score than the proposed Baseline (OT) trained with optimal transport loss.
SHOT achieves the second-highest F1\_score and is better than MVDet, indicating the effectiveness of the multi-height fusion module in SHOT.
MVDeTr achieves the second-highest MODA and the highest MODP among all methods, which proves the advantage of the multi-view deformable fusion layer.
3DROM doesn't perform well because 3DROM is a data augmentation method for handling overfitting issues, but SynMVCrowd is already a large dataset. 
% \red{
SVCW \cite{zhang2024multi} and TrackTacular \cite{Teepe_2024_CVPR} are the latest methods, since the SynMVCrowd dataset contains much more complicated scenarios than other datasets, and their design doesn't achieve satisfying performance on SynMVCrowd. MVOT uses a more precise point-supervision loss instead of the MSE Gaussian density map, and thus achieves higher performance than other methods. But the proposed baseline model still outperforms all comparisons on the  SynMVCrowd dataset, demonstrating the effectiveness of the model under large scenes.
% }
These experiments all indicate that SynMVCrowd is a very challenging platform for multi-view crowd localization evaluation.
See visualization results in Fig. \ref{fig:multiveiw_results}.

% \red{
\subsubsection{Generalization to Novel Scenes} 
We also added generalization experiments to the multi-view crowd localization datasets Wildtrack and MultiviewX, as seen in Table \ref{table:Wildtrack_MultiviewX_results}. In Table \ref{table:Wildtrack_MultiviewX_results}, we compare the results of SOTA multi-view crowd localization methods trained and tested on Wildtrack or MultiviewX solely, with Our Baseline methods' results trained on SynMVCrowd and tested on Wildtrack or MultiviewX. It concludes that our model trained on SynMVCrowd can be well applied to new scenes with simple finetuning, which is fast and achieves comparable performance to SOTAs, eg, MVDet. With the extra domain technique, the performance can be further promoted and outperform MVDet. Even though there is a performance gap between our domain transferring results and single-scene results of the latest SOTAs, such as 3DROM, POV-MVD, etc, SynMVCrowd still provides a new platform for studying better domain transferring methods for the multi-view localization tasks.
% }

\subsection{Multi-view Crowd Counting}

\subsubsection{Methods}
As to multi-view crowd counting, we evaluate and compare the latest SOTA methods on SynMVCrowd, including MVMS \cite{Zhang2020WideAreaCC}, MVMSR \cite{Zhang2020WideAreaCC}, CVCS backbone \cite{zhang2021CVCS}, CVCS \cite{zhang2021CVCS}, and 3DCounting \cite{zhang20203d}. 
% CVCS is a camera selection multi-view counting method and CVCS backbone is the backbone model of CVCS without the camera selection module. 
Besides, we also adopt the proposed baseline method (as in Sec. \ref{subsec:multi-view localization}) for multi-view crowd counting on SynMVCrowd, where a different Gaussian kernel is used to generate ground truth density maps whose sum is equal to the number of persons in the scenario.
Two losses, MSE and OT losses, are also enforced in the baseline model, respectively.
% \textcolor{blue}{But different from the multi-view crowd localization task, we use a different Gaussian kernel to generate ground truth. In the crowd counting task, we need to ensure that the sum of ground truth is equal to the number of persons in the scenario, while in the crowd localization task, there is no such restriction. Therefore, we retrain the proposed model with new ground truth.} \NOTE{(Any difference with the multi-view localization model? The GT is different?)}
% Two losses, MSE and OT losses, are also enforced in the baseline model, respectively.
% Besides, we also show our proposed Baseline (OT)'s (as in Sec. \ref{subsec:multi-view counting and localization}) multi-view crowd counting performance on SynMVCrowd.

\subsubsection{Evaluation Metrics}
For multi-view counting evaluation, we use three metrics:  Mean Absolute Error (MAE), Mean Squared Error (MSE), and mean Normalized Absolute Error (NAE) of the predicted and ground-truth crowd numbers. The formulas are defined as follows: $MAE = \sum (\left | z_{i} - \hat{z_{i}} \right |/ N) $, $ MSE = \sqrt{\sum (z_{i} - \hat{z_{i}})^2 / N } $, $ NAE = \sum \left | z_{i} - \hat{z_{i}} \right | / (z_{i} * N) $, where N is the number of images, $z_{i}$ is the actual number of people existed in the $i$-th image, and $\hat{z_{i}}$ indicates the predicted number of people in the $i$-th image. 

\begin{figure*}[t]
\begin{center}
   \includegraphics[width=\linewidth]{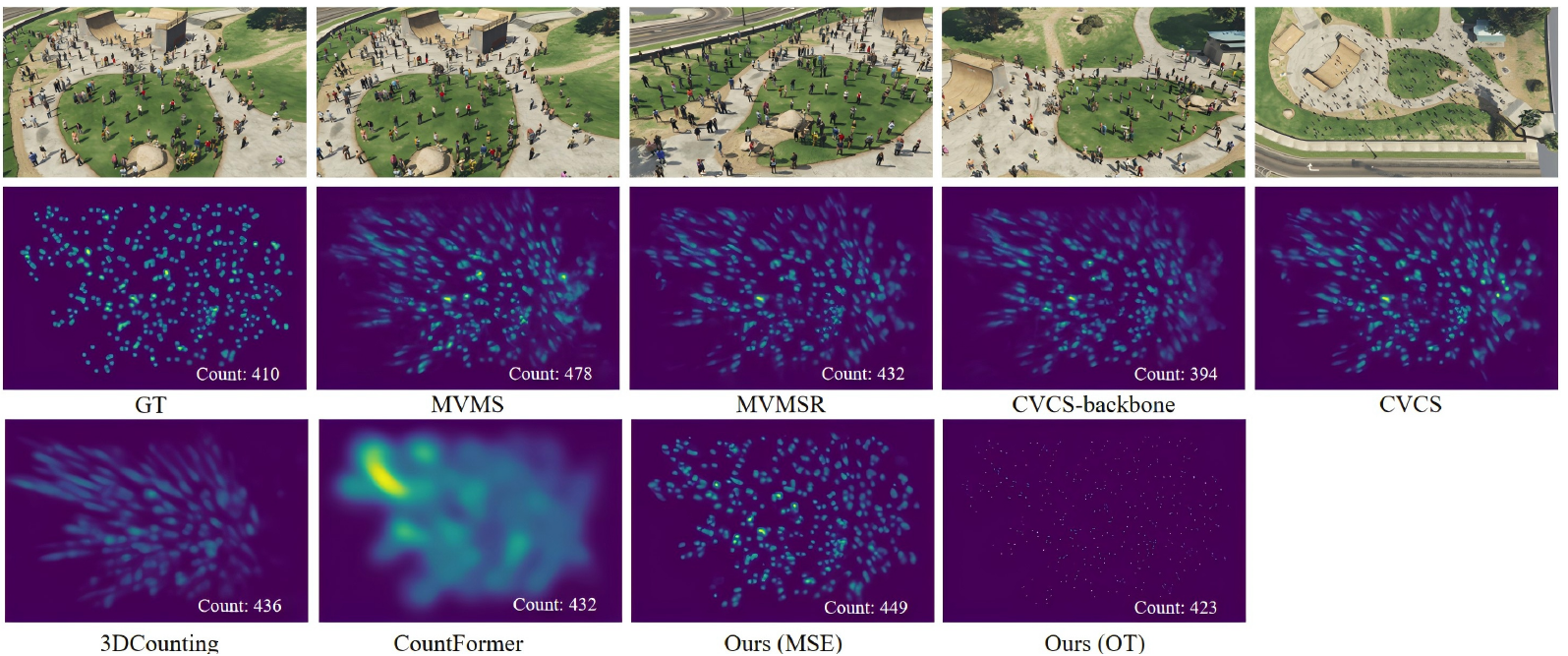}
\end{center}
% \vspace{-0.3cm}
   \caption{The visualization results of multi-view crowd counting: the first row is the camera view input, and the second is the corresponding ground truth (GT) and prediction results.
   }
\vspace{-0.6cm}
\label{fig:mvc_results}
\end{figure*}

\begin{table}[t]
% \small
\centering
\caption{The results of multi-view crowd counting methods on the SynMVCrowd dataset. 
% \vspace{-0.2cm}
% Our baseline method (trained with OT loss) achieves better performance than other methods.
%\NOTE{(Add more multi-view counting methods.
%We may need to rerun all methods to make sure they can work normally.)}
}
\begin{tabular}{lccc}
\toprule
    Method                        & MAE $\downarrow$ & NAE $\downarrow$ & MSE $\downarrow$   \\
\midrule
    MVMS  \cite{Zhang2020WideAreaCC}                               & 81.51  & 0.173  & 102.74  \\
    MVMSR  \cite{Zhang2020WideAreaCC}                              & 70.60  & 0.143  & 93.26  \\
    CVCS backbone \cite{zhang2021CVCS}   & 72.50 & 0.154  & 93.99  \\
    CVCS \cite{zhang2021CVCS}           & 67.74  & 0.139  & 89.31  \\
    3DCounting   \cite{zhang20203d}       & 74.63  & 0.171  & 101.30  \\
    CountFormer \cite{mo2024countformer}   & 56.30 & 0.117  & 79.30  \\
\midrule
    Baseline (MSE)                      & 68.79 & 0.158 & 92.04 \\
    Baseline (OT)                       &  \textbf{50.20}   & \textbf{0.104} & \textbf{68.19}\\

\botrule
\end{tabular}
\vspace{-0.5cm}
\label{table:multi_view_counting_results}
\end{table}

\subsubsection{Experiment Results}
\textbf{Result analysis.}
The results of state-of-the-art multi-view crowd counting methods and the baseline method on SynMVCrowd are shown in Table \ref{table:multi_view_counting_results}, and the visualization results are shown in Fig \ref{fig:mvc_results}. 
MVMS achieves worse performance than MVMSR, showing that the rotation selection layer in MVMSR can effectively alleviate the projection distortion. 
CVCS achieves better performance than the CVCS backbone and MVMSR, indicating the fusion effectiveness of the camera selection module in CVCS. 
%Besides, 3DCounting achieves worst counting performance may caused by the large number of persons in the scene and the more complex occlusion environment, which makes the 3D ground truth cannot enough to provide more information about the characteristics of the obscured person. 
The proposed baseline method `Baseline (OT)' achieves the best performance among all methods.
Overall, all multi-view crowd counting methods generally perform much worse on SynMVCrowd than on other multi-view datasets. Thus, SynMVCrowd is a more difficult benchmark for the multi-view counting task, and more methods that could handle large crowd numbers with more scene and view variations are in demand in this area.

\begin{table}[t]
\centering
    \caption{The multi-view counting and localization performance comparison of using different feature extraction backbones in the Baseline (OT) model, which is trained on the SynMVCrowd dataset with 5 views as input.}
% \label{tab:performance}
\begin{tabular}
{@{\hspace{0.0cm}}l@{\hspace{0.08cm}}c@{\hspace{0.08cm}}c@{\hspace{0.08cm}}c@{\hspace{0.08cm}}|c@{\hspace{0.08cm}}c@{\hspace{0.08cm}}c@{\hspace{0.08cm}}c@{\hspace{0.08cm}}c@{\hspace{0.0cm}}}
% \toprule
\hline
Model & {MAE$\downarrow$} & {MSE$\downarrow$} & {NAE$\downarrow$}  &{MODA$\uparrow$} &{MODP$\uparrow$} &{Precision$\uparrow$} & {Recall$\uparrow$} & {F1\_score$\uparrow$}\\ \hline
VGG19  & 59.04 & 83.00 & 0.155 & 43.8 &69.6 &88.4 &51.6 & 65.0\\ 
ResNet18 & \textbf{50.20} & \textbf{68.19} & \textbf{0.104}& \textbf{49.6} &\textbf{70.2} &\textbf{88.6} &\textbf{57.0} &\textbf{69.4} \\ 
Transformer & 66.23 & 92.84 & 0.211 &39.5  &62.1 &80.1 &48.3 &60.3 \\ \hline
\end{tabular} \label{fig:backbone}
\end{table}

\red{
\begin{table}[t]
\centering
    \caption{Testing the baseline model on the multi-view counting task under different camera view numbers.}
\begin{tabular}{ccccc}
\hline
Models & test views & MAE$\downarrow$ & MSE$\downarrow$ & NAE$\downarrow$ \\ \hline
\multirow{4}{*}{CountFormer} & 3 & 47.71 & 65.24 & 0.151 \\
 & 5 & 56.30 & 79.30 & 0.117 \\
 & 7 & 71.47 & 95.95 & 0.158 \\
 & 9 & 77.82 & 102.96 & 0.164 \\ \hline
\multirow{4}{*}{Baseline (OT)} & 3 & 57.97 & 69.00 & 0.137 \\
 & 5 & 50.20 & 68.19 & 0.104 \\
 & 7 & 46.25 & 59.57 & 0.113 \\
 & 9 & \textbf{42.46} & \textbf{58.34} & \textbf{0.106} \\ \hline
\end{tabular} \label{table: variable views input}
\end{table}
}

% \red{
\textbf{Backbone model ablation study.}
We have conducted an ablation study on the Baseline (OT)'s model architecture by using different single-view feature extraction backbones for the multi-view counting and localization, eg, ResNet18 \cite{he2016deep}, Transformer \cite{liu2022swin}, as in Table \ref{fig:backbone}. 
Each model is trained on SynMVCrowd with 5 views as input, while keeping all other settings the same. From the table, we conclude that the best results are achieved by using ResNet18. The possible reason is that ResNet18 pretrained weights provide more useful information compared to VGG19, and it seems the transformer-based model consists of too many deep layers and cannot perform counting and localization tasks well on the dataset. Thus, we use ResNet18 as the main backbone for the proposed baseline model. 
% }

% \red{
\textbf{Test camera number ablation study.}
We conducted an ablation study on the different numbers of input camera views in Table \ref{table: variable views input}.
The Baseline (OT) and CountFormer model is trained on the SynMVCrowd dataset with 5 input views and is tested with 3-9 camera views in Table \ref{table: variable views input}. We can conclude that the model performance is steadily increased with more camera views, since more camera views can cover the scenes well. In contrast, CountFormer's performance decreases a lot when the camera view number increases, indicating the model is not robust to variable input camera views on the SynMVCrowd dataset, since the multi-view prediction error is accumulated with more cameras. It also demonstrates the proposed baseline model's advantages in fusing multi-camera information under different testing camera views.
% }

\begin{table}[t]
\centering
\caption{
% \red{
Performance Comparison of baseline model with/without spatial feature selection module using MSE or OT loss.}
% }
\label{tab:ablation on spaceselctor}
%\adjustbox{max width=\textwidth}{%
% \begin{tabular}{@{}ll*{8}{S[table-format=3.2]}S[table-format=2.1]@{}}
\begin{tabular}{lcccc|ccccc}
% \toprule
\hline
Loss & Module & MAE$\downarrow$ & MSE$\downarrow$ & NAE$\downarrow$ & MODA$\uparrow$ & MODP$\uparrow$ & Precision$\uparrow$ & Recall$\uparrow$ & F1\_score$\uparrow$ \\
\hline
\multirow{2}{*}{MSE} 
& Without  & 74.33 & 97.52 & 0.236 &31.9 & 74.8 & 94.1 & 34.0 &49.9  \\
& With  & 68.79 & 92.04 & 0.158 &34.6 &74.5 &92.9 &37.4  &53.4  \\
\hline
\multirow{2}{*}{OT}

& Without  & 52.94 & 78.21 & 0.105 & 46.7 & 69.7 & 87.9 & 54.1 & 67.0 \\
& With & \textbf{50.20} & \textbf{68.19} & \textbf{0.104} & \textbf{49.6} & \textbf{70.2} & \textbf{88.6} & \textbf{57.0} & \textbf{69.4} \\
\hline
\end{tabular}%
%}
\end{table}

% \red{
\textbf{Spatial feature selection module ablation study.}
We have conducted an ablation study on the baseline model with/without the spatial module, as shown in Table \ref{tab:ablation on spaceselctor}. We conclude that, no matter using MSE or OT loss, the model with the spatial feature selection outperforms the one without it on both multi-view counting and localization tasks. It demonstrates the effectiveness of the spatial feature selection module for selecting useful features in better multi-view fusion.
% }

\begin{table}[t]
\centering
\small
\caption{
The performance comparison of the different pitch ranges.
}
% \label{tab:performance}
\begin{tabular}{@{\hspace{0.08cm}}l@{\hspace{0.08cm}}c@{\hspace{0.08cm}}c@{\hspace{0.08cm}}c@{\hspace{0.08cm}}|c@{\hspace{0.08cm}}c@{\hspace{0.08cm}}c@{\hspace{0.08cm}}c@{\hspace{0.08cm}}c@{\hspace{0.08cm}}}
\hline
Pitch & {MAE$\downarrow$} & {MSE$\downarrow$} & {NAE$\downarrow$} &{MODA$\uparrow$} &{MODP$\uparrow$} &{Precision$\uparrow$} & {Recall$\uparrow$} & {F1\_score$\uparrow$} \\ 
\hline
$[-90, -60)$ & 48.19& 67.00 & 0.100  &46.0 &64.9 &85.0 & 55.9  &67.4  \\ 
$[-60, -30)$ & 49.62 & 68.29& 0.105 & 45.6 &67.8 &84.8  & 55.6  &67.2\\ 
$[-30, 0]$ & 55.97 & 76.03 & 0.118 & 42.9 &66.7 &83.4  & 53.6  &65.2 \\ 
\hline
\end{tabular} \label{table:pitch}
\end{table}

\begin{table}[t]
\centering
\caption{Computation statistic comparison of different methods. 
%\textit{Note: Memory refers to the model parameter memory size, not the training memory usage.}
}

\label{tab:methods_comparison_on_PFM}
\begin{tabular}{lccc}
\toprule
Methods & Parameters (MB) & FLOPs (G) &Memory (MB)\\
\midrule
MVDet \cite{hou2020multiview} & 30.10 & 910.44 & 114.85 \\
SHOT \cite{song2021stacked} & 19.14 & 1230.24 & 73.04 \\
MVDeTr \cite{hou2021multiview} & 12.78 & 148.30 & 48.78 \\
3DROM \cite{zhang20203d} & 32.46 & 1012.36 & 123.86  \\
SVCW \cite{zhang2024multi} & 25.71 & 3413.13 & 98.12  \\
MVOT \cite{zhang2024mahalanobis}& 18.59 & 349.73 & 70.95 \\
TrackTacular \cite{Teepe_2024_CVPR} & 30.50 & 922.54 & 116.43 \\
Baseline & 18.63 & 349.93 &71.09 \\
\bottomrule
\end{tabular}
\end{table}

% \red{
\textbf{Camera angle influence.} We also provide an ablation study on the camera angle's influence on the crowd counting and localization performance in Table \ref{table:pitch}.
We divide the vertical pitch into three parts for multi-view counting and localization performance analysis: [-90,-60), [-60, -30), and [-30, 0]. Each camera angle group's multi-view crowd counting and localization results are shown in Table \ref{table:pitch}. For each sample in the test set, we randomly select 5 camera views of each frame, where the pitch angle of each view falls within the corresponding angle group.
% where the Baseline (OT) model is tested on each camera angle group with randomly selecting 5 views in each group for 21 times in each frame per scene.
Apparently, the camera pitch significantly influences the performance of multi-view crowd counting and localization. The performance of both counting and detection consistently declines as the camera pitch shifts from downward to frontal views. This trend indicates that frontal perspectives introduce greater occlusion and scale variation, making accurate localization and density estimation more challenging compared to downward-oriented angles.
% }

% \red{
\textbf{Computation statistic comparison.}
Besides, we compare the models' computation statistics, parameter number, FLOPs, and model parameter memory size, as seen in Table \ref{tab:methods_comparison_on_PFM}. Baseline model parameter is smaller than MVDet, due to a smaller feature dimension being used. 3DROM, TrackTacular, and SVCW consume many more parameters because they need a more complicated architecture in the feature fusion or model supervision. It concludes that the baseline model's parameter number is smaller than most methods, indicating its computational efficiency.
% }

% \begin{table}[t]
% % \small
% \centering
% \caption{The multi-view crowd counting domain adaptation results on the CityStreet dataset.}
% % \vspace{-0.3cm}
% \begin{tabular}{l@{\hspace{0.10cm}}l@{\hspace{0.10cm}}c@{\hspace{0.08cm}}c@{\hspace{0.08cm}}c@{\hspace{0.08cm}}}
% \toprule
% Training    &Method                        & MAE $\downarrow$ & NAE $\downarrow$ & MSE $\downarrow$   \\
% \midrule
% \multirow{3}{*}{Single-scene}
%     &MVMS (FCN-7) \cite{zhang2019wide}   & 8.01 & 0.096  & 10.05  \\
%     &MVMS (CSRNet) \cite{Zhang2020WideAreaCC}  & 7.36 & 0.096  & 9.02  \\
%     &MVMSR (CSRNet) \cite{Zhang2020WideAreaCC}   & 6.98 & 0.086  & 8.49  \\
% \midrule
% \multirow{3}{*}{Cross-scene}
%     &Baseline (test)       & 80.92  & 0.944  & 89.43 \\
%     &Baseline (30\%)       & 10.55  & 0.132  & 13.43 \\
%     &Baseline (30\%+DA)         &   \textbf{10.02} & \textbf{0.130} & \textbf{12.16} \\
% \botrule
% \end{tabular}
% \vspace{-0.5cm}
% \label{table:multi_view_counting_da_results}
% \end{table}

\begin{table}[t]
% \small
\centering
\caption{
The multi-view crowd counting domain adaptation (DA) results on the CityStreet dataset and the PETS2009 dataset.}
% \vspace{-0.3cm}
\begin{tabular}{l@{\hspace{0.10cm}}l@{\hspace{0.10cm}}c@{\hspace{0.08cm}}c@{\hspace{0.08cm}}c@{\hspace{0.08cm}}|c@{\hspace{0.08cm}}c@{\hspace{0.08cm}}c@{\hspace{0.08cm}}}
\hline
 & Dataset &  \multicolumn{3}{c|}{CityStreet}  &  \multicolumn{3}{c}{PETS2009}  \\
Training  &Method  & MAE $\downarrow$ & NAE $\downarrow$ & MSE $\downarrow$ & MAE $\downarrow$ & NAE $\downarrow$ & MSE $\downarrow$ \\
\hline
\multirow{10}{*}{Single-scene}
    & Dmap\_wtd \cite{Ryan2014Scene}  & 11.10  & 0.121 & -    & 7.51 & 0.261 & - \\
    & Dect+ReID  \cite{zhang2019wide} & 27.60 & 0.385 & -    & 9.41 & 0.289 & -   \\
    & LateFusion \cite{zhang2019wide} & 8.12 & 0.097  &- & 3.92 & 0.138  &- \\
    & EarlyFusion \cite{zhang2019wide} &8.10 &0.096 & -  & 5.43 &0.199 & -  \\
    & MVMS (FCN-7) \cite{zhang2019wide}   & 8.01 & 0.096  & 10.05   & 3.49  & 0.124  & 4.83   \\
    & MVMS (CSRNet) \cite{Zhang2020WideAreaCC}  & 7.36 & 0.096  & 9.02  & 3.99  & 0.140  & 5.26   \\
    & MVMSR (CSRNet) \cite{Zhang2020WideAreaCC}   & 6.98 & 0.086  & 8.49   & 4.15  & 0.151  & 5.60   \\
    & 3DCounting \cite{zhang20203d} &8.01 &0.096 & -  & 3.15 &0.113 & -  \\
    %& BEVFormer \cite{li2022bevformer} &5.81 &0.078 & -\\
    & CountFormer \cite{mo2024countformer} & 4.72 &  0.058 & -& 0.74 &0.030 & -\\
\hline
\multirow{7}{*}{Cross-scene}
    &Baseline (test)       & 80.92  & 0.944  & 89.43  & 27.92  & 0.868  & 29.10  \\
    % &   &  &  &\\
    % &   &  &  &\\
    % &   &  &  &\\
    &Baseline (10\%)       & 13.95  & 0.155 & 17.37  & 9.60  & 0.324  & 11.43 \\
    &Baseline (30\%)       & 10.55  & 0.132  & 13.43 & 9.37  & 0.317  & 11.16 \\
    &Baseline (100\%)       & 7.24 & 0.085 & 8.96 & 3.46  & 0.116 & 4.57  \\
    &Baseline (10\%+DA)       &   12.67&0.152 & 16.24   & 5.31 & 0.177 &6.46 \\
    &Baseline (30\%+DA)       &   10.02 & 0.130 & 12.16 & 4.26  & 0.147  & 5.49  \\
    &Baseline (100\%+DA)       &  \textbf{6.85} & \textbf{0.083} & \textbf{8.46}  &\textbf{3.29}  & \textbf{0.112} & \textbf{4.54} \\
\hline
\end{tabular}
\vspace{-0.4cm}
\label{table:multi_view_counting_da_CityStreet_PETS2009}
\end{table}

\subsubsection{Generalization to Novel Scenes}
Table \ref{table:multi_view_counting_da_CityStreet_PETS2009} presents the generalization performance of the multi-view counting baseline method trained on SynMVCrowd with OT loss to new scenes CityStreet \cite{zhang2019wide} and PETS2009 \cite{ferryman2009pets2009}. `Baseline (test)' means testing the `Baseline' model (trained on SynMVCrowd) directly on the CityStreet dataset; `Baseline (X\%)' means fine-tuning the `Baseline (OT)' model on the new datasets with $10\%$, $30\%$, and $100\%$ annotations, respectively; `Baseline (X\%+DA)' means adapting the `Baseline (OT)' model to the new datasets with domain adaption approaches and $X\%$ annotations. We train `Baseline (X\%+DA)' with two steps. First, we apply single-image unsupervised domain adaptation to fine-tune the trained model's feature extractor as in \cite{DANN}. In particular, we add a feature discriminator behind the feature extractor of the baseline model to reduce the domain gap between the two datasets. During the fine-tuning stage, both the real and synthetic images are input into the model, and then the features are fed into the discriminator for classification. Next, we input the extracted features into a single-view decoder, expecting that the feature extractor can learn useful feature information. In this process, we only use the annotation of the synthetic dataset. Second, we fix the trained feature extractor and only train the multi-view feature decoder by using X\% CityStreet or PETS2009 annotations and synthetic annotations. 

% \red{
According to Table \ref{table:multi_view_counting_da_CityStreet_PETS2009}, there is a large performance gap between the cross-scene testing `Baseline (test)' and the single-scene methods, such as MVMS \cite{zhang2019wide} or MVMSR \cite{Zhang2020WideAreaCC}, due to the domain gap between the real and synthetic scenes. However, the gap could be largely reduced with a small set of labels for fine-tuning and could be reduced further with domain adaptation techniques (see `Baseline (30\%)' and `Baseline (100\%)', comparing with `Baseline (30\%+DA)' and `Baseline (100\%+DA)').
With fully-supervised finetuning and simple domain techniques, the baseline model trained on SynMVCrowd can also achieve comparable and better performance
compared to 3DCounting and MVMSR, even though still lower than CountFormer.
In the future, the domain adaptation methods will be a key topic in multi-view crowd counting, and our SynMVCrowd benchmark could serve as an important platform.
% }

%Overall, SynMVCrowd is a very challenging and comprehensive platform for multi-view crowd counting and localization tasks. With a large amount of data, SynMVCrowd shall provide new directions and advance the research for multi-view crowd counting and localization more practically. Future demand for new losses and domain-transferring methods may increase with SynMVCrowd.

% \begin{table*}[h]
% % \small
% \centering
% % \begin{tabular}%{c|ccc|c@{\hspace{0.15cm}}c@{\hspace{0.15cm}}c@{\hspace{0.15cm}}c@{\hspace{0.15cm}}c@{\hspace{0.10cm}}}
% % {c|cc|ccc}
% % \hline
% %     Method    & MAE $\downarrow$    & MSE $\downarrow$      & Precision  $\uparrow$      & Recall $\uparrow$    & F1\_score $\uparrow$      \\
% % \hline
% %     CSRNet \cite{li2018csrnet}              &  21.84   &  36.58      & 68.40   & 40.45     & 50.84  \\
% %     DM-Count \cite{wang2020distribution}     &  \textbf{20.88}   &  \textbf{34.72}      & \textbf{98.31}   & \textbf{76.45}     & \textbf{86.01} \\
% %     P2PNet \cite{song2021rethinking}        &  61.74   & 115.03      & 80.59   & 67.62     & 73.54\\
% % \hline
% % \end{tabular}
% \caption{The results of single-image crowd counting and localization on the SynMVCrowd dataset. We report all methods' single-image counting and localization performance.}
% %\vspace{-0.4cm}
% \label{table:single_image_results}
% \end{table*}

\subsection{Single-image Counting and Localization}

\subsubsection{Methods}

In addition to multi-view crowd counting and localization, SynMVCrowd could also serve as a challenging platform for evaluating and comparing single-image counting and localization tasks. 
% Single-image crowd counting and localization aims to estimate the number of people in an input image of an unconstrained scene and map an input crowd image to its corresponding density map which indicates the number of people per pixel present. 
%For the single-image crowd localization task, we estimate locations through one-to-one matching using the Hungarian algorithm, evaluated by computation of Precision, Recall, and F1\_score at fixed distance thresholds. 
%All the formulas are described in Sec \ref{subsec:multi-view counting and localization}.
We test and compare 7 mainstream single-image counting and localization methods with publicly available codes on SynMVCrowd:
CSRNet \cite{li2018csrnet}, DM-Count \cite{wang2020distribution}, NoisyCC \cite{wan2020Modeling}, GLoss \cite{Wan2021CVPR} and P2PNet \cite{song2021rethinking}, 
CLTR \cite{liang2022end}, STEERER \cite{haniccvsteerer}, GramFormer \cite{lin2024gramformer} and FreeLunch \cite{meng2025free}.
Besides, we use domain adaptation techniques to improve the baseline counting model's cross-scene generalization performance trained on SynMVCrowd and compare it with the performance of single-scene methods.
We use MAE, MSE, and NAE to evaluate the counting performance and Precision, Recall, and F1\_score to evaluate the localization performance (as in the metrics of Sec. \ref{subsec:multi-view localization}).

% \textbf{Experiment methods.}
% We test and compare five mainstream single-image counting and localization methods on SynMVCrowd:
% CSRNet \cite{li2018csrnet}, DM-Count \cite{wang2020distribution}, NoisyCC \cite{wan2020Modeling}, GLoss \cite{Wan2021CVPR} and P2PNet \cite{song2021rethinking}.

%In single-image counting and localization task, we select five mainstream methods trained on our proposed dataset: 
% CSRNet \cite{li2018csrnet} is a classical and efficient crowd counter in which the author designs a dilatation module and adds it to the top of the backbone; DM-Count \cite{wang2020distribution} uses optimized transport to evaluate the similarity between normalized predictions and the ground truth. Compared with the Gaussian smoothing method, the proposed method has more strict model loss control conditions; NoisyCC \cite{wan2020Modeling} proposes a novel loss function that explicitly models annotation noise as random variables. The proposed loss function will reduce the weight of the uncertain region, making it robust to the tagging noise and improving the training of the density map estimator;
% GLoss \cite{Wan2021CVPR} investigates learning the density map representation through an unbalanced optimal transport problem, and proposes a generalized loss function to learn density map for crowd counting and localization; }
% P2PNet \cite{song2021rethinking} performs crowd-counting tasks through a one-to-one matching strategy based on the Hungarian algorithm. This network significantly improves performance in crowd counting.

\subsubsection{Experiment settings}
 For SynMVCrowd, the training set, the validation set, and the test set are also divided in a 3:1:1 ratio. During training, we randomly select 10 frames from the 200 frames of each scene per epoch for training. For evaluation, we select all frames of each scene in the test set. In the experiments, all the models are trained using the official codes and the default parameters. Besides, we randomly crop patches as inputs from images with a size of $512\times512$ as a way of data augmentation. The distance threshold for evaluating localization performance is 3 pixels for all methods. 
 %In CSRNet \cite{li2018csrnet},
 The ground truth density map for crowd counting is generated by a Gaussian kernel with a fixed size of 3. All experiments are conducted on one RTX3090 GPU.

% \textbf{Baseline and results.}
% \zq{Describe the baseline method.}

\begin{table}[t]
%\small
\centering
% \vspace{-0.3cm}
\caption{The performance of single-image crowd counting and localization SOTA methods (\textbf{with code available}) on the SynMVCrowd dataset. We report all methods' single-image counting and localization performance.
}
\begin{tabular}{@{\hspace{0.cm}}l@{\hspace{0.15cm}}c@{\hspace{0.15cm}}c@{\hspace{0.15cm}}c@{\hspace{0.15cm}}c@{\hspace{0.15cm}}c@{\hspace{0.10cm}}}
%\begin{tabular}{c|cc|ccc}
\toprule
    Method    & MAE$\downarrow$    & MSE$\downarrow$      & Precision$\uparrow$      & Recall$\uparrow$    & F1\_score$\uparrow$      \\
\midrule
    CSRNet \cite{li2018csrnet}              &  21.84   &  36.58      & 50.77   & 69.59     & 58.71  \\
    DM-Count \cite{wang2020distribution}     &  \textbf{20.88}   &  \textbf{34.72}      & 65.53   & 67.31     & 66.41 \\
    NoisyCC  \cite{wan2020Modeling}       &  26.19   & 43.23      & 33.65   & 48.28     & 39.66\\
    P2PNet \cite{song2021rethinking}        &  61.74   & 115.03      & 80.59   & 67.62     & 73.54\\
    GLoss \cite{Wan2021CVPR}         &  21.37   & 40.36      & 79.37   & 70.63     & 74.75\\
    CLTR \cite{liang2022end}  & 23.86 & 38.53 & 85.47 & 66.02 & 74.50 \\
    STEERER \cite{haniccvsteerer}  & 22.81 & 41.75 & \textbf{88.83} & \textbf{67.79} & \textbf{76.70} \\
    GramFormer \cite{lin2024gramformer}  & 31.38 & 50.85 & 22.11 & 23.60 &21.83  \\
    FreeLunch \cite{meng2025free} & 32.56 & 43.40 & 19.68 & 23.56 & 18.18 \\
\botrule
\end{tabular}
% \vspace{-0.2cm}
\label{table:single_image_results}
\end{table}

\begin{table}[t]
%\small
\centering
% \vspace{-0.3cm}
\caption{The results on detailed evaluation granularity. We divide the scenes into 3 types regarding crowd numbers: `Sparse (0-399)', `Medium (400-699)', and `Congested (700-1000)', and evaluate methods accordingly.
%\NOTE{(Add descriptions to the paper.)}
}
\begin{tabular}{lcccccc}
\toprule
    \multirow{2}{*}{Method} & \multicolumn{2}{c}{Sparse (0-399)} & \multicolumn{2}{c}{Medium (400-699)} & \multicolumn{2}{c@{\hspace{0cm}}}{Congested (700-1000)} \\
    & MAE $\downarrow$ & MSE $\downarrow$         & MAE $\downarrow$ & MSE $\downarrow$       & MAE $\downarrow$ & MSE $\downarrow$          \\
\midrule
CSRNet  \cite{li2018csrnet}    & \textbf{14.92}       & \textbf{23.69}  & 27.06    & 42.75       &  45.17    & 64.85 \\ 
DMCount     \cite{wang2020distribution} &18.23          &29.76   & \textbf{25.00}      &  \textbf{41.00}         & \textbf{35.34}       & \textbf{56.58}  \\ 
NoisyCC  \cite{wan2020Modeling}      & 20.45       & 31.01  & 29.78    & 43.77       & 47.61     & 68.95 \\ 
P2PNet  \cite{song2021rethinking}  & 27.77          & 35.15       & 45.26    & 66.01         & 79.96     & 92.04  \\         
GLoss  \cite{Wan2021CVPR}  & 17.68       & 27.26  & 26.49    & 42.33       &  37.74    & 57.37 \\ 
CLTR   \cite{liang2022end}& 18.22      & 28.91  & 30.91    & 47.64       &  45.23    & 64.14 \\ 
STEERER  \cite{haniccvsteerer}  & 17.12       & 30.10  & 31.06    & 47.67       &  52.91    & 72.57 \\ 
GramFormer \cite{lin2024gramformer}  & 24.38 & 36.98 & 41.34 & 68.77 & 81.35 & 125.93\\
FreeLunch \cite{meng2025free}  & 28.96 & 38.30 & 41.08 & 54.79 & 66.79 & 82.93 \\
\botrule
\end{tabular}

% \vspace{-0.6cm}
\label{table:granularity_results}
\end{table}

\begin{figure*}[t]
\begin{center}
\includegraphics[width=\linewidth]{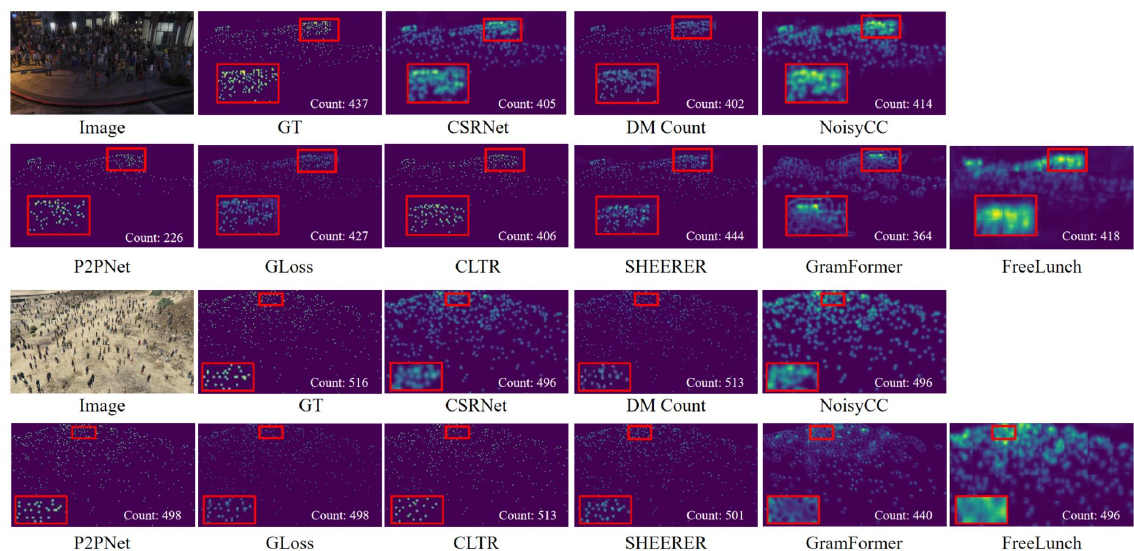}
\end{center}
% \vspace{-0.1cm}
\caption{The visualization results of single-image crowd counting and localization.}
%\NOTE{(Add more vis examples of single-image counting and localization. Also draw a figure of domain adaption counting results on SHA and SHAB.)}
%}
% \vspace{-0.3cm}
\label{fig:single_image_results}
\end{figure*}

\subsubsection{Experiment results}
\textbf{Result analysis.}
We compare single-image crowd counting and localization methods on SynMVCrowd in Table \ref{table:single_image_results}. 
DM-Count performs the best among all methods in terms of crowd counting performance, and GLoss achieves slightly worse counting results than DM-Count, showing that the optimal transport method is quite robust for single-image crowd counting in different environments. 
As to single-image localization, STEERER \cite{haniccvsteerer} is the best one among all methods, and CLTR \cite{liang2022end} achieves the second-best localization results. Compared to other density map regression methods (such as CSRNet \cite{li2018csrnet}, DM-Count \cite{wang2020distribution}, NoisyCC \cite{wan2020Modeling}), point supervision methods can achieve better localization results. 
%DM-Count performs the best among all methods in terms of both crowd counting and localization performance, and GLoss achieves the second-best counting result, showing that the optimal transport method is quite robust for single-image crowd counting in different environments.  In the single-image localization task, GLoss is the best one among all methods and DM-Count achieves slightly worse localization results than GLoss. 
In particular, NoisyCC achieves the worst localization results and the second-worst counting performance. 
The possible reason is that the annotation noise in SynMVCrowd is rare due to no human labeling process, which reduces NoisyCC's noise modeling effectiveness.
%modeling the annotation noise as a random variable as in NoisyCC is not very effective in the large synthetic dataset, whose annotation noise is relatively small without human labeling efforts, resulting in a performance decline of this method in counting and localization.
CSRNet achieves slightly worse counting results than DM-Count and GLoss, but CSRNet achieves much worse localization results, because DM-Count and GLoss use optimal transport with better localization ability than the density maps supervision of CSRNet.

Surprisingly, P2PNet achieves the worst counting results and the third-best localization performance. The reason may be that P2PNet is a point proposal prediction method, and its prediction results are more likely to be affected by factors such as light and camera placement angle changes. 
% \red{
In the case of the latest works, GramFormer \cite{lin2024gramformer} and FreeLunch \cite{meng2025free}, both fail to deliver promising results. GramFormer requires controlling the head width of datasets, while FreeLunch relies on thermal maps as ground truth. Since SynMVCrowd does not meet these specific requirements, we set the head width to a fixed value of 5 pixels and omitted the corresponding thermal map loss. This is the primary reason why the results of these methods are relatively worse compared to others.
% }
In SynMVCrowd, we have more horizontal camera views and around $50,000$  out of 5.3 million images under low illumination conditions, whose numbers are more than any existing single-image dataset. Thus, it is a rather challenging dataset. 
Fig. \ref{fig:single_image_results} demonstrates the visualization results of all comparisons.

%Besides, CSRNet achieves better counting results possibly because it uses Gaussian density maps as supervision so that it can achieve better counting performance in the case of losing certain localization accuracy compared to P2PNet, CLTR and STEERER. 
% Thus, P2PNet shows worse counting results but better localization performance than CSRNet.
%Besides, CSRNet achieves better counting results possibly because it uses Gaussian fuzzy operation so that it can achieve better counting performance in the case of losing certain localization accuracy. 
% Compared with the localization performance of DM-Count and GLoss, it proves that the optimal transport method is quite robust for single-image crowd localization in different environments.
% The reason is P2PNet is a method for localization, whose counting performance is evaluated by the number of true positives and many people are missed. Still, P2PNet achieves better localization performance than CSRNet, but is worse than DM-Count.

% In a word, in addition to multi-view counting and localization tasks, SynMVCrowd can also serve as a useful and challenging benchmark for single-image counting and localization tasks.

\textbf{Count granularity.} Besides, we also evaluate all methods of higher granularity in Table \ref{table:granularity_results}. In detail, we divide the scenes into 3 types regarding crowd numbers: Sparse (0-399), Medium (400-699), and Congested (700-1000), and evaluate methods accordingly. It shows P2PNet \cite{song2021rethinking} performs badly in Congested scenes, where SynMVCrowd contains many scenes with large crowd numbers (700-1000); Even though CSRNet \cite{li2018csrnet} is old, it is good on sparse scenes; DMCount \cite{wang2020distribution} is better in congested scenes, thus it shows the best counting performance.

\subsubsection{Generalization to Novel Scenes} 
\begin{figure}[t]
\begin{center}
%\vspace{-0.4cm}
   % \includegraphics[width=1\linewidth]{figures/SynMVCrowd Single-imgae Domain Adaptation Introduction.pdf}
   \includegraphics[width=1\linewidth]{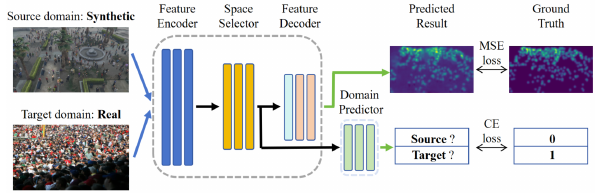}
\end{center}
%\vspace{-0.2cm}
   %\caption{The dataset samples of different scale variations: `Large', `Medium' and `Small'.}
  \caption{Architecture of the domain adaptation network for single-image crowd counting. The model takes synthetic (source) and real (target) images as input. Features are encoded and selected, then processed through two branches: 1) A decoder producing density maps, supervised by MSE loss against ground truth; 2) A domain predictor distinguishing source from target features using cross-entropy (CE) loss. This adversarial training aligns feature distributions across domains to improve generalization.}
\vspace{-0.2cm} \label{fig: single-image domain adaptation}
\end{figure}

To evaluate the effectiveness of our proposed dataset, we conduct experiments to present the generalization performance of the baseline counting method trained on SynMVCrowd to novel scenes. 
% \red{ 
The single-image domain transferring model's detailed structure is shown in Fig. \ref{fig: single-image domain adaptation} and Table \ref{table:layer_setting}.

The single-image feature extractor is based on VGG19, and the single-image decoder is the same as in DM-Count \cite{wang2020distribution}. When transferring the baseline to novel new scenes, we randomly crop patches as inputs from images with a size of $512 \times 512$. After the single-image feature extractor, we obtain the image features of size [1, 512, 64, 64], which contains an up-sampling operation with the size of two scale factors. In discrimination, we first transform the input image features into features of size [1,8,64,64] through the convolutional layer. Next, the features are converted into a two-dimensional vector of $[1, 8\times64\times64]$, which is then input into the fully connected layer to obtain category predictions.

\begin{SCtable}
\small
\centering
%\begin{tabular}{ll}
\caption {The single-image baseline for domain adaptation. The single-image baseline includes a single-image feature extractor, a single-image decoder, and a discriminator. The Filter dimensions are input channels, output channels, and filter size ($w\!  \times\!  h$).}
\footnotesize
\begin{tabular}{|c|c|}
\hline
\multicolumn{2}{|c|}{Single-image baseline } \\ \hline
\multicolumn{2}{|c|}{Single-image feature extractor}  \\ \hline
%\multicolumn{2}{|c|}{\centering \textcolor{blue}{VGG19 \cite{simonyan2014very}, ResNet18 {\cite{he2016deep}}}} \\ \hline

% Layer & Filter     \\ \hline
% conv 1     & $3\!  \times\!  64\!  \times\!  3\!  \times\!  3$, relu   \\ %\hline
% conv 2     & $64\!  \times\!  64\!  \times\!  3\!  \times\!  3$, relu, maxpooling  \\ %\hline
% conv 3     & $64\!  \times\!  128\!  \times\!  3\!  \times\!  3$, relu   \\ %\hline
% conv 4     & $128\!  \times\!  128\!  \times\!  3\!  \times\!  3$, relu, maxpooling   \\ %\hline
% conv 5     & $128\!  \times\!  256\!  \times\!  3\!  \times\!  3$, relu  \\ %\hline
% conv 6     & $256\!  \times\!  256\!  \times\!  3\!  \times\!  3$, relu  \\ %\hline
% conv 7     & $256\!  \times\!  256\!  \times\!  3\!  \times\!  3$, relu  \\ %\hline
% conv 8     & $256\!  \times\!  256\!  \times\!  3\!  \times\!  3$, relu, maxpooling   \\ %\hline
% conv 9     & $256\!  \times\!  512\!  \times\!  3\!  \times\!  3$, relu  \\ %\hline
% conv 10     & $512\!  \times\!  512\!  \times\!  3\!  \times\!  3$, relu  \\ %\hline
% conv 11     & $512\!  \times\!  512\!  \times\!  3\!  \times\!  3$, relu  \\ %\hline
% conv 12     & $512\!  \times\!  512\!  \times\!  3\!  \times\!  3$, relu, maxpooling  \\ %\hline
% conv 13     & $512\!  \times\!  512\!  \times\!  3\!  \times\!  3$, relu   \\ 
% conv 14     & $512\!  \times\!  512\!  \times\!  3\!  \times\!  3$, relu   \\
% conv 15     & $512\!  \times\!  512\!  \times\!  3\!  \times\!  3$, relu   \\
% conv 16     & $512\!  \times\!  512\!  \times\!  3\!  \times\!  3$   \\ \hline
Layer & Feature Extractor \\ \hline
Layer & UpSampling \\ \hline
\multicolumn{2}{|c|}{Single-image decoder}\\ \hline
Layer & Filter \\ \hline
conv 1      & $512\!  \times\!  256\!  \times\!  3\!  \times\!  3$,   relu \\
conv 2      & $256\!  \times\!  128\!  \times\!  3\!  \times\!  3$, relu \\
conv 3      & $128\! \times\!  1\!   \times\! 1\!    \times\! 1$    \\  \hline

\multicolumn{2}{|c|}{Discriminator}  \\ \hline
Layer & Filter     \\ \hline
% conv 1     & $512\!  \times\!  512\!  \times\!  3\!  \times\!  3$, relu   \\ %\hline
% conv 2     & $512\!  \times\!  512\!  \times\!  3\!  \times\!  3$, relu   \\ %\hline
% conv 3     & $1\!  \times\!  512\!  \times\!  3\!  \times\!  1$   \\ \hline
conv  & $512\!  \times\!  8\! \times\!1 \times\!1$ \\  
linear1  & $32768\!  \times\!  1024\! $ , relu\\
linear2 & $1024\!  \times\!  1024\! $ , relu\\
linear3 & $1024\!  \times\!  2\! $\\  \hline
\end{tabular}
\vspace{0.1cm}
\label{table:layer_setting}
\centering
\vspace{-0.5cm}
\end{SCtable}

Table \ref{table:single_image_counting_da_results} reports the results of the single-scene methods \cite{MCNN,li2018csrnet,wang2020distribution,Wan2021CVPR,Cheng_2022_CVPR} and our cross-scene methods on ShanghaiTech A and ShanghaiTech B \cite{zhang2016single}. `Test (GCC)' means testing the baseline counting model trained on GCC dataset (a synthetic single-image counting dataset \cite{Wang2019Learning}) directly on ShanghaiTech A and ShanghaiTech B; `Test (SynMVCrowd)' means testing the baseline model trained on our proposed SynMVCrowd dataset directly on above two datasets; `SE Cycle GAN (GCC)' represents the SE Cycle GAN method in \cite{Wang2019Learning} trained on GCC for domain adaptation to above two datasets; `UDA (SynMVCrowd)' means adapting the baseline model trained on SynMVCrowd to above two datasets with a unsupervised domain adaption technique, which is as follows. 

For `UDA (SynMVCrowd)', we use a domain adversarial method \cite{DANN} to reduce the domain gap between the two different datasets. The baseline counting model contains a backbone module to extract multi-scale features and a density map regression head.`UDA (SynMVCrowd)' contains an extra classification head to distinguish synthetic and real images, based on the baseline counting model. During the training process, all dataset images are fed into the backbone module, and then we get the extracted features. Next, we input the features into the regression head, expecting to obtain a reliable single-view crowd density prediction by only using synthetic annotations for supervision. In addition, we use the classification head to classify features to encourage the backbone module to learn domain-invariant feature information from the different domains.
% Specifically, we use mean teacher module \cite{Tarvainen_Valpola_2017} to reduce the domain gap between different datasets. The baseline model contains a backbone model to extract multi-scale features, a regression head and a classification head. We construct the regression head and classification head for density estimation and class prediction, respectively. And the overall framework is based on the traditional mean teacher framework where the student and teacher share the same architecture and the teacher is updated with the exponential moving average (EMA) weights of the student. The synthetic images and annotations are used to train the student model by leveraging the supervised loss. The unlabeled real samples are exploited for encouraging model to transfer domain style. We deliberately blurred the image, such as masking, data enhancement, \textit{etc}, and fed it into the student model, while still asking it to make predictions about these obscured area. In order to achieve the purpose of supervision, we use the teacher model to predict the origin images and generate pseudo labels to supervise the prediction results of the student model. %

\begin{table}[t]
% \small
\centering
% \vspace{-0.2cm}
\caption{The single-image crowd counting domain adaptation results on the ShanghaiTech A and B dataset.
%\NOTE{(Update with Supplemental results. Update corresponding paragraphs.)}
}
\vspace{-0.2cm}
\begin{tabular}{l@{\hspace{0.10cm}}l@{\hspace{0.10cm}}c@{\hspace{0.08cm}}c@{\hspace{0.08cm}}c@{\hspace{0.08cm}}c@{\hspace{0.08cm}}}
\toprule
\multirow{2}{*}{Training}   &Dataset          & \multicolumn{2}{c}{SHA}    & \multicolumn{2}{c}{SHB} \\
&Method                        & MAE $\downarrow$ & MSE $\downarrow$  & MAE $\downarrow$ & MSE $\downarrow$ \\
\midrule
\multirow{8}{*}{Single-scene}
    & MCNN \cite{MCNN} & 110.20 & 173.20 & 26.40 & 41.30 \\
    & CSRNet \cite{li2018csrnet} & 68.20 & 115.00 & 10.60 & 16.00 \\
    & DM-Count \cite{wang2020distribution} & 59.70 & 95.70 & 7.40 & 11.80 \\
    & GLoss \cite{Wan2021CVPR} & 61.30 & 95.40 & 7.30 & 11.70 \\
    & GauNet \cite{Cheng_2022_CVPR} & 54.80 & 89.10 & 6.20 & 9.90 \\
    &PET \cite{liu2023point} & 49.3 & 78.70 & 6.10 & 9.60  \\
    &APGCC \cite{chen2024improving} & 48.80 & 76.70 & 5.60 & 8.70 \\
    &DiffusionLoc \cite{zhang2025diffusionloc} & 50.30 & 79.10 & 5.40 & 8.50 \\
\midrule
\multirow{4}{*}{Cross-scene}
    & Test (GCC) & 160.80 & 216.50 & 22.80 & 30.60 \\
    & Test (SynMVCrowd) & 187.70 & 291.90 & 18.10 & 28.30 \\
    & SE Cycle GAN (GCC) & 123.40 & 193.40 & 19.90 & 28.30 \\
    & UDA (SynMVCrowd) & \textbf{118.70} & \textbf{182.90} & \textbf{11.20} & \textbf{18.40} \\
\botrule
\end{tabular}
\vspace{-0.5cm}
\label{table:single_image_counting_da_results}
\end{table}

\begin{figure*}[t]
\begin{center}
\includegraphics[width=\linewidth]{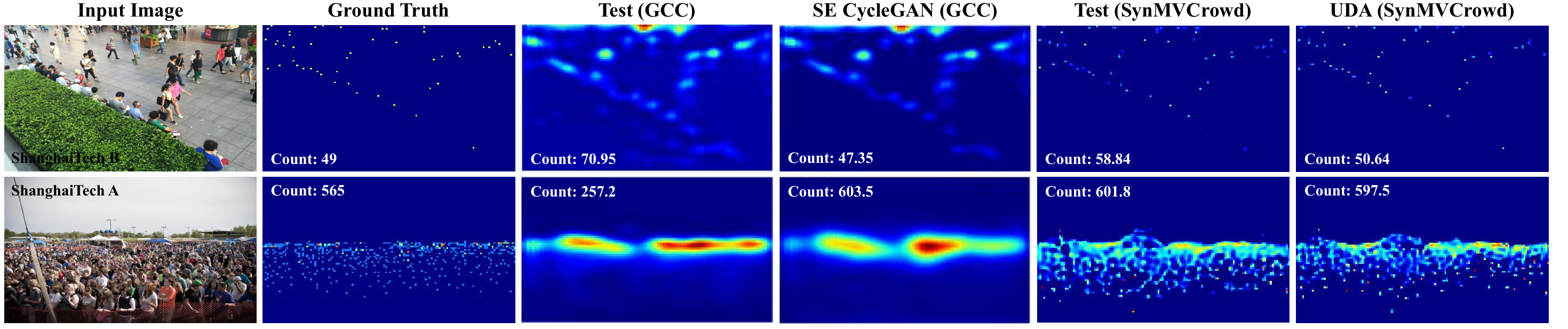}
\end{center}
\vspace{-0.1cm}
\caption{The domain adaptation visualization results of single-image crowd counting on ShanghaiTech B and ShanghaiTech A.}
%). In first row, we show the visualization result test on ShanghaiTech B. In second row, we show the visualization result test on ShanghaiTech A.}
\vspace{-0.5cm}
\label{fig:single_image_da}
\end{figure*}

\begin{table}[t]
\small
\centering
\caption{
% \red{
The single-image crowd counting domain adaptation results on UCF-CC-50 and Mall datasets. Comparisons are all trained and tested on the same datasets, while ours is trained on SynMVCrowd and tested on UCF-CC-50 or Mall.}
% }
\label{tab:dmcount synmvcrowd to mall}
%\resizebox{0.8\textwidth}{!}{ % 调整表格宽度为文本宽度的90%
\begin{tabular}{lccc|ccc}
% \toprule
\hline
Datasets &\multicolumn{3}{c}{UCF-CC-50} &\multicolumn{3}{c}{MALL} \\ 
\hline
Method & MAE$\downarrow$& NAE$\downarrow$  &MSE$\downarrow$ & MAE$\downarrow$ & NAE$\downarrow$  & MSE$\downarrow$ \\ \hline
Crowd-CNN \cite{crowd-cnn2015} & 466.90 & - & 498.60 & 1.83 & - & 2.76 \\
MCNN \cite{MCNN}            & 377.60 & - & 509.10 & 2.74 & - & 13.50 \\
CSRNet \cite{li2018csrnet}  & 266.10 & - & 397.50 & - & - & - \\
DecideNet \cite{decidenet2018} & 361.80 & - & 493.50 & 1.53 & - & 1.89 \\
IG-CNN \cite{ig-cnn2018} & 419.60 & - & 541.80 & 2.39 & - & 9.14 \\
ST-CNN \cite{st-cnn2019} & 229.40 & - & 325.60 & 4.03 & - & 5.87 \\
DM-Count \cite{wang2020distribution} & 211.00 & - & 291.50 & - & - & - \\
P2PNet \cite{song2021rethinking} & 172.72 & - & 256.18 & - & - & - \\
U-ASD Net \cite{uasdnet2021} & 232.30 & - & 217.80 & 1.80 & - & 2.20 \\
GRGAF-ST \cite{grgaf2023} & - & - & - & 1.61 & - & 2.07 \\
PET \cite{liu2023point} & 159.90& - & 223.70 & - & - & - \\
CL-DCNN \cite{cl-dcnn2024} & 181.80 & - & 240.60 & 1.55 & - & 2.01 \\
APGCC \cite{chen2024improving} & 154.80 & - & 205.50 & - & - & - \\
DiffusionLoc \cite{zhang2025diffusionloc} & 135.30 & - & 201.70 & - & - & - \\
\hline
Baseline(test) &  1052.04 & 0.763 & 1367.31 & 21.67 & 0.729 & 21.95\\
Baseline(10\%) & 278.52 & 0.271 & 339.40 & 1.99 & 0.065& 2.55 \\
Baseline(30\%) & 266.28 & 0.260& 329.19 & 1.67 & 0.055 & 2.13 \\
Baseline(100\%) &249.67 & 0.250 & 315.52 & 1.39 & 0.046 & 1.81 \\
Baseline(10\%+DA) &229.82 & 0.237& 315.49 & 1.84 & 0.061 & 2.33 \\
Baseline(30\%+DA) &214.57 & 0.231 & 284.40 & 1.41 & 0.046 & 1.80 \\
Baseline(100\%+DA) & \textbf{202.34} &\textbf{0.224} & \textbf{265.14} & \textbf{1.34} & \textbf{0.044} & \textbf{1.74} \\
\hline
\end{tabular}
%}
\end{table}

According to Table \ref{table:single_image_counting_da_results}, due to the domain gap between real and synthetic scenarios, there is a gap among the performance of the cross-scene test (`Test (GCC)',  `Test (SynMVCrowd)') and the single-scene SOTA methods. However, this gap can be reduced by unsupervised domain adaptation techniques, as in `UDA (SynMVCrowd)'. The visualization results are shown in Fig. \ref{fig:single_image_da}. 
% Besides, `Test (SynMVCrowd)’ is better than `Test (GCC)’ and `SE Cycle GAN (GCC)', because the larger number of images in our dataset can cover more scenarios in practice.
From the results, `Test (SynMVCrowd)' achieves a better counting performance than `Test (GCC)'. Besides, `UDA (SynMVCrowd)' achieves a much lower MAE metric score of 11.2 compared to `SE Cycle GAN (GCC)', which scores 19.9.
In addition, we present the generalization results on ShanghaiTech A (more crowds) in Table \ref{table:single_image_counting_da_results} left. 
We conclude that the model trained on our dataset with a simple unsupervised method (UDA) could achieve better performance compared to utilizing a specially designed domain transferring method, SE Cycle GAN trained on the GCC dataset. % It also proves the effectiveness of the proposed dataset for the crowd-counting task.}
Both the above results demonstrate that the larger number of images in our dataset can cover more scenarios in practice.
% demonstrating that the proposed benchmark has better generalization ability compared to the GCC dataset, whether under the direct test or a domain adaptation approach.
In conclusion, SynMVCrowd can advance related research and serve as a challenging benchmark for single-image counting and localization tasks.

% \red{
Besides, we also perform the single-image counting domain transferring experiments on UCF\_CC\_50 \cite{6619173} and Mall \cite{chen2012feature}, whose results are shown in Table \ref{tab:dmcount synmvcrowd to mall}.
The model trained on SynMVCrowd cannot perform well by directly testing on novel new scenes UCF\_CC\_50 and Mall datasets, since the domain gap between SynMVCrowd and the two datasets is quite large. But, with more labeled target dataset annotations for finetuning, the performance on both datasets increased steadily, and with 100\% finetuning annotations, our domain transferring results already outperform existing SOTA methods on Mall. With further domain adaptation techniques (100\%+DA), the performance can be further improved.
On UCF\_CC\_50, since the crowd count in the images varies from tens to thousands, there is a much larger domain gap with SynMVCrowd, and thus the performance is lower than recent single-scene SOTAs, but we still achieve better performance than DM-Count and U-ASD Net. 
Since SynMVCrowd contains various scenes, camera views, and environmental conditions, it can serve as an important benchmark for single-image counting and domain transferring tasks.
% }

\section{Discussion and Conclusion}
In this paper, we propose a synthetic crowd benchmark called SynMVCrowd for multi-view and single-image crowd counting and localization tasks.
As far as we know, SynMVCrowd is the largest synthetic dataset for multi-view and single-image crowd vision tasks, which consists of a large number of crowd scenes, camera views, and frames with a large crowd number in each scene. 
Several state-of-the-art multi-view and single-image crowd counting and localization methods are evaluated and compared on the benchmark. Besides, we propose strong baselines and achieve better performance than comparisons, and the experiments show SynMVCrowd can be applied to novel scenes with domain adaptation techniques. It proves that SynMVCrowd can serve as a challenging platform for more practical research for multi-view and single-image crowd counting and localization.
In addition to multi-view and single-image counting and localization tasks, the proposed dataset can also be used in tasks like crowd segmentation, 2D/3D pose estimation, camera calibrations, novel view synthesis, or large-scene 3D reconstruction, \textit{etc}.

\section*{Acknowledgements}
This work was supported in parts by Guangdong S\&T Program (2024B0101050004), NSFC (62202312), ICFCRT (W2441020), Shenzhen Science and Technology Program (KJZD20240903100022028, KQTD20210811090044003, RCJC20200714114435012), and Scientific Foundation for Youth Scholars of Shenzhen University.

\section*{Declarations}

\textbf{Ethical consideration}.  We generate a synthetic multi-view image dataset as a challenging platform for multi-view and single-image counting and localization, with no real persons recorded or used in the model training and testing, which shall reduce the ethical issues for the public caused by crowd counting and localization tasks.

\textbf{Data availability}.  The data supporting this study's findings is available from the Visual Computing Research Center, Shenzhen University. The dataset is available from the authors upon reasonable request and with permission from the Visual Computing Research Center.
% \bibliographystyle{sn-jnl}
% \bibliography{egbib}
\vfill

\end{document}